\documentclass[a4paper,12pt]{article}

\usepackage{comment}
\usepackage{fancyvrb}
\usepackage[round]{natbib}
\usepackage{graphicx}
\usepackage{hyperref}
\usepackage[ragged]{footmisc}
\usepackage{ragged2e}
\usepackage[section]{placeins}

\title{The SP theory of intelligence: an overview\footnote{Now published as {\em The SP theory of intelligence: an overview} (J G Wolff, {\em Information}, 4 (3), 283-341, 2013, doi:10.3390/info4030283).}}

\author{J Gerard Wolff\footnote{Dr Gerry Wolff, BA (Cantab), PhD (Wales), CEng, MBCS (CITP); CognitionResearch.org, Menai Bridge, UK; \href{mailto:jgw@cognitionresearch.org}{jgw@cognitionresearch.org}; +44 (0) 1248 712962; +44 (0) 7746 290775; {\em Skype}: gerry.wolff; {\em Web}: \href{http://www.cognitionresearch.org}{www.cognitionresearch.org}.}}

\normalsize

\date{September 2013}

\begin{document}

\maketitle

\begin{abstract}

This article is an overview of the {\em SP theory of intelligence}, which aims to simplify and integrate concepts across artificial intelligence, mainstream computing and human perception and cognition, with information compression as a unifying theme. It is conceived as a brain-like system that receives `New' information and stores some or all of it in compressed form as `Old' information; and it is realised in the form of a computer model, a first version of the {\em SP machine}. The matching and unification of patterns and the concept of {\em multiple alignment} are central ideas. Using heuristic techniques, the system builds multiple alignments that are `good' in terms of information compression. For each multiple alignment, probabilities  may be calculated for associated inferences. Unsupervised learning is done by deriving new structures from partial matches between patterns and via heuristic search for sets of structures that are `good' in terms of information compression. These are normally ones that people judge to be `natural', in accordance with the `DONSVIC' principle---the discovery of natural structures via information compression. The SP theory provides an interpretation for concepts and phenomena in several other areas including `computing', aspects of mathematics and logic, the representation of knowledge, natural language processing, pattern recognition, several kinds of reasoning, information storage and retrieval, planning and problem solving, information compression, neuroscience, and human perception and cognition. Examples include the parsing and production of language with discontinuous dependencies in syntax, pattern recognition at multiple levels of abstraction and its integration with part-whole relations, nonmonotonic reasoning and reasoning with default values, reasoning in Bayesian networks including `explaining away', causal diagnosis, and the solving of a geometric analogy problem.

\end{abstract}


\section{Introduction}\label{introduction_section}

The {\em SP theory of intelligence}, which has been under development since about 1987,\footnote{Apart from the period between early 2006 and late 2012 when I was working on other things.} aims to simplify and integrate concepts across artificial intelligence, mainstream computing and human perception and cognition, with information compression as a unifying theme.

The name `SP' is short for {\em Simplicity} and {\em Power}, because compression of any given body of information, $I$, may be seen as a process of reducing informational `redundancy' in $I$ and thus increasing its `simplicity', whilst retaining as much as possible of its non-redundant expressive `power'. Likewise with Occam's Razor (Section \ref{occams_razor_section}, below).

Aspects of the theory, as it has developed, have been described in several peer-reviewed articles.\footnote{See \href{http://www.cognitionresearch.org/sp.htm\#PUBS}{www.cognitionresearch.org/sp.htm\#PUBS}.} The most comprehensive description of the theory as it stands now, with many examples, is in \citet{wolff_2006}.

But this book, with more than 450 pages, is too long to serve as an introduction to the theory. This article aims to meet that need, with a fairly full description of the theory and a selection of examples.\footnote{Some of the text and figures in this article come from the book, with permission. Details of other permissions are given at appropriate points in the article.} For the sake of brevity, the book will be referred to as `{\em BK}'.

The next section describes the origins and motivation for the SP theory, Section \ref{introduction_to_sp_theory_section} introduces the theory, Sections \ref{multiple_alignment_section} and \ref{unsupervised_learning_section} fill in a lot of the details, while the following sections describe aspects of the theory and what it can do.

\section{Origins and motivation}\label{origins_motivation_section}

The following subsections outline the origins of the SP theory, how it relates to some other research, and how it has developed.

\subsection{Information compression}\label{information_compression_section}

Much of the inspiration for the SP theory is a body of research, pioneered by Fred Attneave \citeyearpar{attneave_1954}, Horace Barlow \citeyearpar{barlow_1959,barlow_1969}, and others, showing that several aspects of the workings of brains and nervous systems may be understood in terms of information compression.\footnote{Also relevant and still of interest is Zipf's \citeyearpar{zipf_1949} {\em Human Behaviour and the Principle of Least Effort}. Incidentally, Barlow later suggested that ``... the [original] idea was right in drawing attention to the importance of redundancy in sensory messages ... but it was wrong in emphasizing the main technical use for redundancy, which is compressive coding.''  \citep[][p. 242]{barlow_2001_network}. As we shall see, the SP theory is closer to Barlow's original thinking than what he said later.} For example, when we view a scene with two eyes, the image on the retina of the left eye is almost exactly the same as the image on the retina of right eye, but our brains merge the two images into a single percept, and thus compress the information \citep{barlow_1969}.\footnote{This focus on compression of information in binocular vision is distinct from the more usual interest in the way that slight differences between the two images enables us to see the scene in depth.}

More immediately, the theory has grown out of my own research, developing models of the unsupervised learning of a first language, where the importance of information compression became increasingly clear \citep[eg,][]{wolff_1988}.\footnote{Details of other relevant publications may be found via \href{http://bit.ly/12DOkTV}{bit.ly/12DOkTV}.}

The theory also draws on principles of `minimum length encoding' pioneered by \citet{solomonoff_1964}, and others. And it has become apparent that several aspects of computing, mathematics, and logic may be understood in terms of information compression ({\em BK}, Chapters 2 and 10).

At an abstract level, information compression can bring two main benefits:

\begin{itemize}

\item For any given body of information, $I$, information compression may reduce its size and thus facilitate the storage, processing and transmission of $I$.

\item Perhaps more important is the close connection between information compression and concepts of prediction and probability \citep[see, for example,][]{li_vitanyi_2009}. In the SP system, it is the basis for all kinds of inference, and for calculations of probabilities.

\end{itemize}

In animals, we would expect these things to have been favoured by natural selection because of the competitive advantage they can bring. Notwithstanding the `QWERTY' phenomenon,\footnote{The dominance of the QWERTY keyboard despite its known inefficiencies.} there is reason to believe that information compression, properly applied, may yield comparable advantages in artificial systems.

\subsection{The matching and unification of patterns}\label{matching_and_unification_section}

In the SP theory, the matching and unification of patterns is seen as being closer to the bedrock of information compression than more mathematical techniques such as wavelets or arithmetic coding, and closer to the bedrock of information processing and intelligence than, say, concepts of probability. A working hypothesis in this programme of research is that, by staying close to relatively simple, `primitive', concepts of matching patterns and unifying them, there is a better chance of cutting through unnecessary complexity, and in gaining new insights and better solutions to problems. The mathematical basis of wavelets, arithmetic coding, and probabilities, may itself be founded on the matching and unification of patterns ({\em BK}, Chapter 10).

\subsection{Simplification and integration of concepts}\label{occams_razor_section}

In accordance with Occam's Razor, the SP system aims to combine conceptual simplicity with descriptive and explanatory power. Apart from this widely-accepted principle, the drive for simplification and integration of concepts in this research programme has been motivated in part by Allen Newell's critique of some kinds of research in cognitive science \citep{newell_1973}, and in part by the apparent fragmentation of research in artificial intelligence and mainstream computing, with their myriad of concepts and many specialisms.

In attempting to simplify and integrate ideas, the SP theory belongs in the same tradition as unified theories of cognition such as Soar \citep[][]{laird_2012} and ACT-R \citep[][]{anderson_etal_2004}.\footnote{Both of them inspired by Allen Newell \citep[eg,][]{newell_1973}.} And it chimes with the resurgence of interest in understanding artificial intelligence as a whole \citep[see, for example,][]{agi_2011} and with research on `natural computation' \citep[][]{dodig-crnkovic_2011}.

Although the SP programme shares some objectives with projects such as the G{\"o}del Machine \citep[][]{steunebrink_schmidhuber_2011}, and `universal artificial intelligence' \citep[][]{hutter_2005}, the approach is very different.

\subsection{Transparency in the representation of knowledge}\label{transparency_in_kr_section}

In this research, it is assumed that knowledge in the SP system should normally be transparent or comprehensible, much as in the `symbolic' tradition in artificial intelligence (see also Section \ref{donsvic_section}), and distinct from the kind of `sub-symbolic' representation of knowledge that is the rule in, for example, `neural networks' as they are generally conceived in computer science.

As we shall see in Section \ref{representation_of_knowledge_section} and elsewhere in this article, SP patterns in the multiple alignment framework may serve to represent a variety of kinds of knowledge, in symbolic forms.

\subsection{Development of the theory}

In developing the theory, it was apparent at an early stage that existing systems---such my models of language learning\footnote{See \href{http://www.cognitionresearch.org/lang\_learn.html}{www.cognitionresearch.org/lang\_learn.html}.} and systems like Prolog---would need radical re-thinking to meet the goal of simplifying and integrating ideas across a wide area.

The first published version of the SP theory \citep{wolff_1990} described `some unifying ideas in computing'. Early work on the SP computer model concentrated on developing an improved version of the `dynamic programming' technique for the alignment of two sequences (see {\em BK}, Appendix A) as a possible route to modelling human-like flexibility in pattern recognition, analysis of language, and the like.

About 1992, it became apparent that the explanatory range of the theory could be greatly expanded by forming alignments of 2, 3, or more sequences, much as in the `multiple alignment' concept of bioinformatics. That idea was developed and adapted in new versions of the SP model, and incorporated in new procedures for unsupervised learning.

Aspects of the theory, with many examples, have been developed in \citet{wolff_2006}.

\section{Introduction to the SP theory}\label{introduction_to_sp_theory_section}

\begin{figure}[!htbp]
\centering
\includegraphics[width=0.5\textwidth]{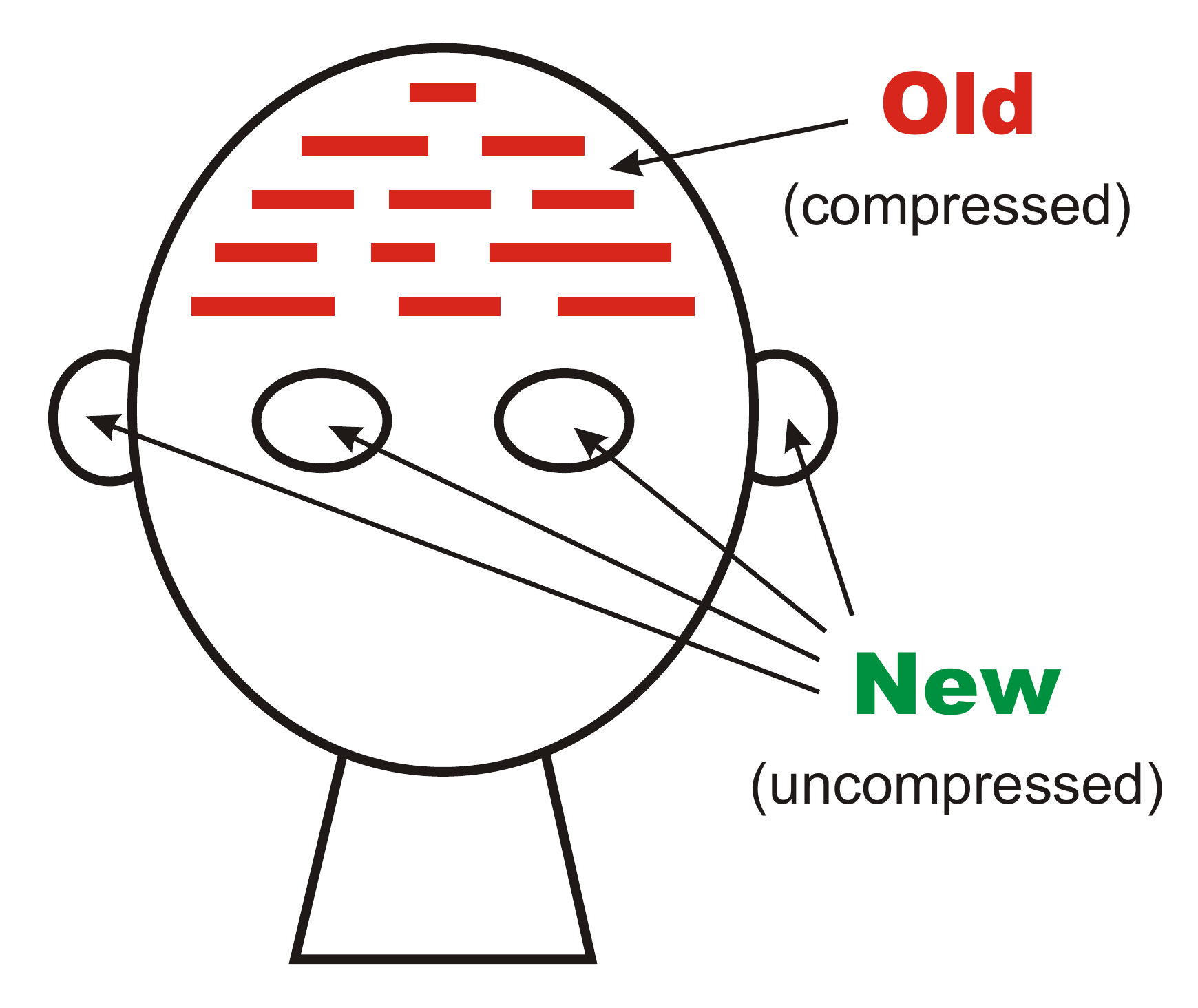}
\caption{Schematic representation of the SP system from an `input' perspective.}
\label{sp_input_perspective_figure}
\end{figure}

The main elements of the SP theory are:

\begin{itemize}

\item The SP theory is conceived as an abstract brain-like system that, in an `input' perspective, may receive `New' information via its senses, and store some or all of it in its memory as `Old' information, as illustrated schematically in Figure \ref{sp_input_perspective_figure}. There is also an `output' perspective, described in Section \ref{decompression_by_compression_section}.

\item The theory is realised in the form of a computer model, introduced in Section \ref{computer_model_section}, below, and described more fully later.

\item All New and Old information is expressed as arrays ({\em patterns}) of atomic symbols in one or two dimensions. An example of an SP pattern may be seen in each row in Figure \ref{parsing_1_figure}. Each symbol can be matched in an all-or-nothing manner with any other symbol. Any meaning that is associated with an atomic symbol or group of symbols must be expressed in the form of other atomic symbols.

\item Each pattern has an associated frequency of occurrence which may be assigned by the user or derived via the processes for unsupervised learning. The default value for the frequency of any pattern is 1.

\item The system is designed for the unsupervised learning of Old patterns by compression of New patterns.\footnote{Of course, people can and do learn with assistance from teachers and others. But unsupervised learning has been a focus of interest in developing the SP theory, since it is clear that much of our learning is done without assistance, and because unsupervised learning raises some interesting issues and yields some useful insights, as outlined in Section \ref{donsvic_section}.}

\item An important part of this process is, where possible, the economical (compressed) encoding of New patterns in terms of Old patterns. This may be seen to achieve such things as pattern recognition, parsing or understanding of natural language, or other kinds of interpretation of incoming information in terms of stored knowledge, including several kinds of reasoning.

\item In keeping with the remarks in Section \ref{matching_and_unification_section}, compression of information is achieved via the matching and unification (merging) of patterns. In this, there are key roles for the frequency of occurrence of patterns, and their sizes.

\item The concept of {\em multiple alignment}, described in Section \ref{multiple_alignment_section}, is a powerful central idea, similar to the concept of multiple alignment in bioinformatics but with important differences.

\item Owing to the intimate connection, previously mentioned, between information compression and concepts of prediction and probability, it is relatively straightforward for the SP system to calculate probabilities for inferences made by the system, and probabilities for parsings, recognition of patterns, and so on (Section \ref{ma_probabilities_section}).

\item In developing the theory, I have tried to take advantage of what is known about the psychological and neurophysiological aspects of human perception and cognition, and to ensure that the theory is compatible with such knowledge (see Section \ref{perception_cognition_neuroscience_section}).

\end{itemize}

\subsection{The SP computer model}\label{computer_model_section}

The SP theory is realised most fully in the SP70 computer model, with capabilities in the building of multiple alignments and in unsupervised learning.\footnote{At the time of creating this arXiv version of the article, the latest version of the SP computer model is SP71. The source code for SP71, with a Windows executable file and some other files, are in the ancillary files in arXiv for this version of the article.} This will be referred to as the SP model, although in some cases examples are from a subset of the model or slightly earlier precursors of it.

The SP model and its precursors have played a key part in the development of the theory:

\begin{itemize}

\item As an antidote to vagueness. As with all computer programs, processes must be defined with sufficient detail to ensure that the program actually works.

\item By providing a convenient means of encoding the simple but important mathematics that underpins the SP theory, and performing relevant calculations, including calculations of probability.

\item By providing a means of seeing quickly the strengths and weaknesses of proposed mechanisms or processes. Many ideas that looked promising have been dropped as a result of this kind of testing.

\item By providing a means of demonstrating what can be achieved with the theory.

\end{itemize}

The workings of the SP model is described in some detail in {\em BK} (Sections 3.9, 3.10, and 9.2) and more briefly in Sections \ref{multiple_alignment_section} and \ref{unsupervised_learning_section}, below. The source code for the models, with associated documents and files, may be downloaded via links under the heading `SOURCE CODE' at the bottom of the page on \href{http://bit.ly/WtXa3g}{bit.ly/WtXa3g}.

The two main elements of the models, described in the following two sections, are the building of multiple alignments and the unsupervised learning of Old patterns.

\subsection{The SP machine}\label{sp_machine_section}

The SP model may be regarded as a first version of the {\em SP machine}, an expression of the SP theory and a means for it to be applied.

\begin{figure}[!htbp]
\centering
\includegraphics[width=0.9\textwidth]{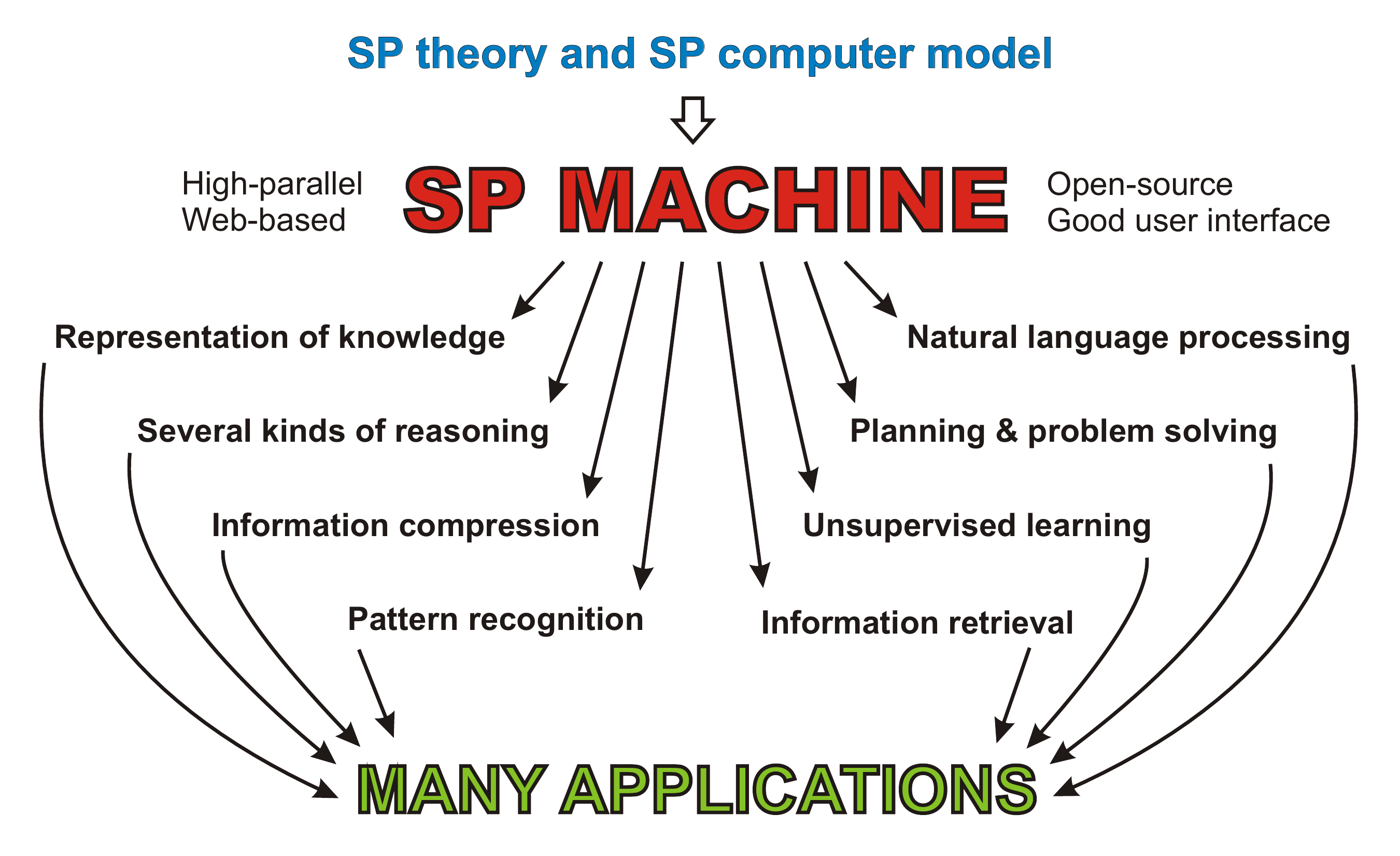}
\caption{Schematic representation of the development and application of the proposed SP machine.}
\label{sp_machine_figure}
\end{figure}

A useful step forward in the development of the SP theory would be the creation of a high-parallel, open-source version of the SP machine, accessible via the web, and with a good user interface.\footnote{As in ordinary search engines, and, indeed, in the brains of people and other animals, high levels of parallelism are needed to achieve speedy processing with large data sets.}  This would provide a means for researchers to explore what can be done with the system and to improve it. How things may develop is shown schematically in Figure \ref{sp_machine_figure}.

The high-parallel search mechanisms in any of the existing internet search engines would probably provide a good foundation for the proposed development.

Further ahead, there may be a case for the creation of new kinds of hardware, dedicated to the building of multiple alignments and other processes in the SP framework \citep[][Section 6.13]{sp_benefits_apps}.

\subsection{Unfinished business}\label{unfinished_business_section}

Like most theories, the SP theory has shortcomings, but it appears that they may be overcome. At present, the most immediate problems are:

\begin{itemize}

\item {\em Processing of information in two or more dimensions}. No attempt has yet been made to generalise the SP model to work with patterns in two dimensions, although that appears to be feasible to do, as outlined in {\em BK} (Section 13.2.1). As noted in {\em BK} (Section 13.2.2), it is possible that information with dimensions higher than two may be encoded in terms of patterns in one or two dimensions, somewhat in the manner of architects' drawings. A 3D structure may be stitched together from several partially-overlapping 2D views, in much the same way that, in digital photography, a panoramic view may be created from partially-overlapping pictures \citep[][Sections 6.1 and 6.2]{sp_vision}.

\item {\em Recognition of perceptual features in speech and visual images}. For the SP system to be effective in the processing of speech or visual images, it seems likely that some kind of preliminary processing will be required to identify low level perceptual features such as, in the case of speech, phonemes, formant ratios, or formant transitions, or, in the case of visual images, edges, angles, colours, luminances, or textures. In vision, at least, it seems likely that the SP framework itself will prove relevant since edges may be seen as zones of non-redundant information between uniform areas containing more redundancy and, likewise, angles may be seen to provide significant information where straight edges, with more redundancy, come together \citep[][Section 3]{sp_vision}. As a stop-gap solution, the preliminary processing may be done using existing techniques for the identification of low-level perceptual features \citep[][Chapter 13]{prince_2012}.

\item {\em Unsupervised learning}. A limitation of the SP computer model as it is now is that it cannot learn intermediate levels of abstraction in grammars (eg, phrases and clauses), and it cannot learn the kinds of discontinuous dependencies in natural language syntax that are described in Sections \ref{discontinuous_dependencies_section} to \ref{aux_verb_2_section}. I believe these problems are soluble and that solving them will greatly enhance the capabilities of the system for the unsupervised learning of structure in data (Section \ref{unsupervised_learning_section}).

\item {\em Processing of numbers}. The SP model works with atomic symbols such as ASCII characters or strings of characters with no intrinsic meaning. In itself, the SP system does not recognise the arithmetic meaning of numbers such as `37' or `652' and will not process them correctly. However, the system has the potential to handle mathematical concepts if it is supplied with patterns representing Peano's axioms or similar information ({\em BK}, Chapter 10). As a stop-gap solution in the SP machine, existing technologies may provide whatever arithmetic processing may be required.

\end{itemize}

\section{The multiple alignment concept}\label{multiple_alignment_section}

The concept of {\em multiple alignment} in the SP theory has been adapted from a similar concept in bioinformatics, where it means a process
of arranging, in rows or columns, two or more DNA sequences or amino-acid sequences so that matching symbols---as many as possible---are aligned orthogonally in columns or rows.

Multiple alignments like these are normally used in the computational analysis of (symbolic representations of) sequences of DNA bases or sequences of amino acid residues as part of the process of elucidating the structure, functions or evolution of the corresponding molecules. An example of this kind of multiple alignment is shown in Figure \ref{DNA_figure}.

\begin{figure}[!htbp]
\fontsize{10.00pt}{12.00pt}
\centering
{\bf
\begin{BVerbatim}
  G G A     G     C A G G G A G G A     T G     G   G G A
  | | |     |     | | | | | | | | |     | |     |   | | |
  G G | G   G C C C A G G G A G G A     | G G C G   G G A
  | | |     | | | | | | | | | | | |     | |     |   | | |
A | G A C T G C C C A G G G | G G | G C T G     G A | G A
  | | |           | | | | | | | | |   |   |     |   | | |
  G G A A         | A G G G A G G A   | A G     G   G G A
  | |   |         | | | | | | | |     |   |     |   | | |
  G G C A         C A G G G A G G     C   G     G   G G A
\end{BVerbatim}
}
\caption{A `good' alignment amongst five DNA sequences.}
\label{DNA_figure}
\end{figure}

As in bioinformatics, a multiple alignment in the SP system is an arrangement of two or more patterns in rows (or columns), with one pattern in each row (or column).\footnote{Whether multiple alignments are shown with patterns in rows or in columns depends largely on what fits best on the page.} The main difference between the two concepts is that, in bioinformatics, all sequences have the same status, whereas in the SP theory, the system attempts to create a multiple alignment which enables one New pattern (sometimes more) to be encoded economically in terms of one or more Old patterns. Other differences are described in {\em BK} (Section 3.4.1).

In Figure \ref{parsing_1_figure}, row 0 contains a New pattern representing a sentence: `\texttt{t h i s b o y l o v e s t h a t g i r l}', while each of rows 1 to 8 contains an Old pattern representing a grammatical rule or a word with grammatical markers. This multiple alignment, which achieves the effect of parsing the sentence in terms of grammatical structures, is the best of several built by the model when it is supplied with the New pattern and a set of Old patterns that includes those shown in the figure and several others as well.

In this example, and others in this article, `best' means that the multiple alignment in the figure is the one that enables the New pattern to be encoded most economically in terms of the Old patterns, as described in Section \ref{ma_evaluation_section}, below.

\begin{figure}[!htbp]
\fontsize{07.00pt}{08.40pt}
\centering
{\bf
\begin{BVerbatim}
0          t h i s        b o y            l o v e s           t h a t        g i r l           0
           | | | |        | | |            | | | | |           | | | |        | | | |
1          | | | |        | | |            | | | | |           | | | |    N 0 g i r l #N        1
           | | | |        | | |            | | | | |           | | | |    |           |
2          | | | |        | | |            | | | | |    NP D   | | | | #D N           #N #NP    2
           | | | |        | | |            | | | | |    |  |   | | | | |                  |
3          | | | |        | | |            | | | | |    |  D 1 t h a t #D                 |     3
           | | | |        | | |            | | | | |    |                                 |
4          | | | |        | | |        V 0 l o v e s #V |                                 |     4
           | | | |        | | |        |             |  |                                 |
5 S NP     | | | |        | | |    #NP V             #V NP                               #NP #S 5
    |      | | | |        | | |     |
6   |  D 0 t h i s #D     | | |     |                                                           6
    |  |           |      | | |     |
7   NP D           #D N   | | | #N #NP                                                          7
                      |   | | | |
8                     N 1 b o y #N                                                              8
\end{BVerbatim}
}
\caption{The best multiple alignment found by the SP model with the New pattern `\texttt{t h i s b o y l o v e s t h a t g i r l}' and a set of user-supplied Old patterns representing some of the grammatical forms of English, including words with their grammatical markers.}
\label{parsing_1_figure}
\end{figure}

\subsection{Coding and the evaluation of an alignment in terms of compression}\label{ma_evaluation_section}

This section describes in outline how multiple alignments are evaluated in the SP model. More detail may be found in {\em BK} (Section 3.5).

Each Old pattern in the SP system contains one or more `identification' symbols or {\em ID-symbols} which, as their name suggests, serve to identify the pattern. Examples of ID-symbols in Figure \ref{parsing_1_figure} are `\texttt{D}' and `\texttt{0}' at the beginning of `\texttt{D 0 t h i s \#D}' (row 6), and `\texttt{N}' and `\texttt{1}' at the beginning of `\texttt{N 1 b o y \#N}' (row 8).

Associated with each type of symbol (where a `type' of symbol is any one of a set of symbols that match each other exactly) is a notional {\em code} or bit pattern that serves to distinguish the given type from all the others. This is only notional because the bit patterns are not actually constructed. All that is needed for the purpose of evaluating multiple alignments is the size of the notional bit pattern associated with each type. This is calculated via the Shannon-Fano-Elias coding scheme (described in \citet{cover_thomas_1991}), using information about the frequency of occurrence of each Old pattern, so that the shortest codes represent the most frequent symbol types and {\em vice versa}.\footnote{Although this scheme is slightly less efficient than the well-known Huffman scheme, it has been adopted because, unlike the Huffman scheme, it does not produce anomalous results when probabilities are derived from code sizes, as described in {\em BK} (Section 3.7).} Notice that these bit patterns and their sizes are totally independent of the names for symbols that are used in written accounts like this one: names that are chosen purely for their mnemonic value.

Given a multiple alignment like the one shown in Figure \ref{parsing_1_figure}, one can derive a {\em code pattern} from the multiple alignment in the following way:

\begin{enumerate}

\item Scan the multiple alignment from left to right looking for columns that contain an ID-symbol by itself, not aligned with any other symbol.

\item Copy these symbols into a code pattern in the same order that they appear in the multiple alignment.

\end{enumerate}

\noindent The code pattern derived in this way from the multiple alignment shown in Figure \ref{parsing_1_figure} is `\texttt{S 0 1 0 1 0 \#S}'. This is, in effect, a compressed representation of those symbols in the New pattern that are aligned with Old symbols in the multiple alignment. In this case, the code pattern is a compressed representation of {\em all} the symbols in the New pattern but it often happens that some of the symbols in the New pattern are not matched with any Old symbols and then the code pattern will represent only those New symbols that are aligned with Old symbols.

In the context of natural language processing, it perhaps more plausible to suppose that the encoding of a sentence is some kind of representation of the meaning of the sentence, instead of a pattern like `\texttt{S 0 1 0 1 0 \#S}'. How a meaning may be derived from a sentence via multiple alignment is described in {\em BK} (Section 5.7).

\subsubsection{Compression difference and compression ratio}

Given a code pattern like `\texttt{S 0 1 0 1 0 \#S}', we may calculate a `compression difference'\index{CD} as:

\begin{equation}
CD = B_N - B_E
\label{CD_equation}
\end{equation}

\noindent or a `compression ratio'\index{CR} as:

\begin{equation}
CR = B_N / B_E,
\label{CR_equation}
\end{equation}

\noindent where $B_N$ is the total number of bits in those symbols in the New pattern that are aligned with Old symbols in the alignment and $B_E$ is the total number of bits in the symbols in the code pattern, and the number of bits for each symbol is calculated via the Shannon-Fano-Elias scheme as mentioned above.

$CD$ and $CR$ are each an indication of how effectively the New pattern (or those parts of the New pattern that are aligned with symbols within Old patterns in the alignment) may be compressed in terms of the Old patterns that appear in the given multiple alignment. The $CD$ of a multiple alignment---which has been found to be more useful than $CR$---may be referred to as the {\em compression score} of the multiple alignment.

In each of these equations, $B_N$ is calculated as:

\begin{equation}
B_N = \sum_{i=1}^h C_i,
\label{BN_equation}
\end{equation}

\noindent where $C_i$ is the size of the code for $i$th symbol in a sequence, $H_1 ... H_h$, comprising those symbols within the New pattern that are aligned with Old symbols within the multiple alignment.

$B_E$ is calculated as:

\begin{equation}
B_E = \sum_{i=1}^s C_i,
\label{BE_equation}
\end{equation}

\noindent where $C_i$ is the size of the code for $i$th symbol in the sequence of $s$ symbols in the code pattern derived from the multiple alignment.

\subsection{The building of multiple alignments}\label{building_multiple_alignments_section}

This section describes in outline how the SP model builds multiple alignments. More detail may be found in {\em BK} (Section 3.10).

Multiple alignments are built in stages, with pairwise matching and alignment of patterns. At each stage, any partially-constructed multiple alignment may be processed as if it was a basic pattern and carried forward to later stages. This is broadly similar to some programs for the creation of multiple alignments in bioinformatics.\footnote{See, for example, ``Sequence alignment'', {\em Wikipedia}, \href{http://en.wikipedia.org/wiki/Sequence\_alignment}{en.wikipedia.org/wiki/Sequence\_alignment}, retrieved 2013-05-08.} At all stages, the aim is to encode New information economically in terms of Old information and to weed out multiple alignments that score poorly in that regard.

The model may create Old patterns for itself, as described in Section \ref{unsupervised_learning_section}, but when the formation of multiple alignments is the focus of interest, Old patterns may be supplied by the user. In all cases, New patterns must be supplied by the user.

At each stage of building multiple alignments, the operations are as follows:

\begin{enumerate}

\item Identify a set of `driving' patterns and a set of `target' patterns. At the beginning, the New pattern is the sole driving pattern and the Old patterns are the target patterns. In all subsequent stages, the best of the multiple alignments formed so far (in terms of their $CD$ scores) are chosen to be driving patterns and the target patterns are the Old patterns together with a selection of the best multiple alignments formed so far, including all of those that are driving patterns.

\item Compare each driving pattern with each of the target patterns to find full matches and good partial matches between patterns. This is done with a process that is essentially a form of `dynamic programming' \citep[][]{sankoff_kruskall_1983}, somewhat like the WinMerge utility for finding similarities and differences between files.\footnote{See \href{http://winmerge.org/}{winmerge.org}.} The process is described quite fully in {\em BK} (Appendix A) and outlined in Section \ref{matching_section}, below. The main difference between the SP process and others, is that the former can deliver several alternative matches between a pair of patterns, while WinMerge and standard methods for finding alignments deliver one `best' result.

\item From the best of the matches found in the current stage, create corresponding multiple alignments and add them to the repository of multiple alignments created by the program.

\end{enumerate}

This process of matching driving patterns against target patterns and building multiple alignments is repeated until no more multiple alignments can be found. For the best of the multiple alignments created since the start of processing, probabilities are calculated, as described in Section \ref{ma_probabilities_section}.

\subsubsection{Finding good matches between patterns}\label{matching_section}

\begin{figure}[!htbp]
\begin{center}
\includegraphics[width=0.7\textwidth]{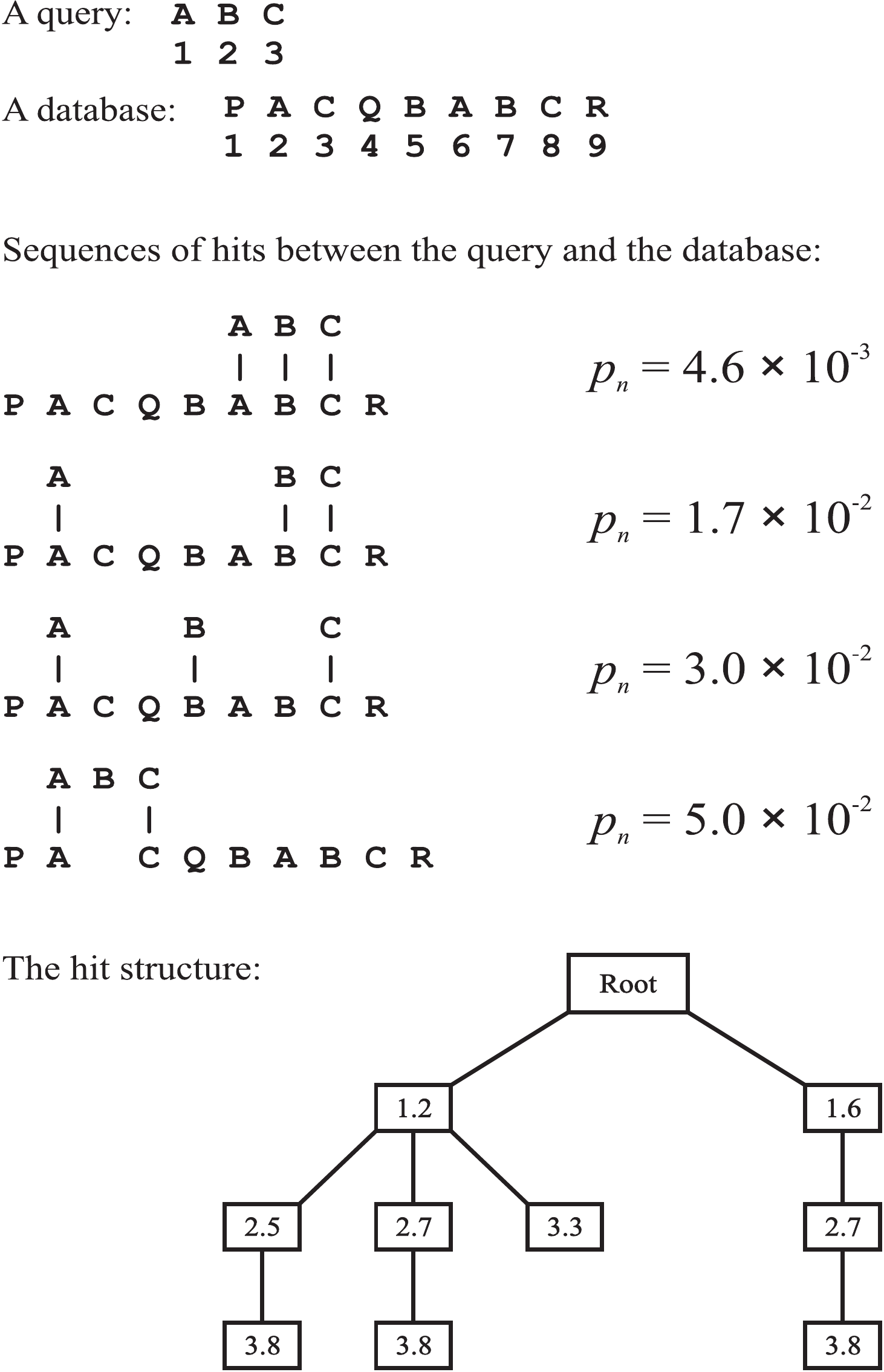}
\end{center}
\caption{An example to show how the SP model finds good full and partial matches between patterns. A `query' string and a `database' string are shown at the top with the ordinal positions of symbols marked. Sequences of hits between the query and the database are shown in the middle with corresponding values of $p_n$ (described in {\em BK}, Section A.2). Each node in the hit structure shows the ordinal position of a query symbol and the ordinal position of a matching database symbol. Each path from the root node to a leaf node represents a sequence of hits.}
\label{pattern_matching_figure}
\end{figure}

Figure \ref{pattern_matching_figure} shows with a simple example how the SP model finds good full and partial matches between a `query' string of atomic symbols (alphabetic characters in this example) and a `database' string:

\begin{enumerate}

\item The query is processed left to right, one symbol at a time.

\item Each symbol in the query is, in effect, broadcast to every symbol in the database to make a yes/no match in each case.

\item Every positive match (hit) between a symbol from the query and a symbol in the database is recorded in a {\em hit structure}, illustrated in the figure.

\item If the memory space allocated to the hit structure is exhausted at any time then the hit structure is purged: the leaf nodes of the tree are sorted in reverse order of their probability values and each leaf node in the bottom half of the set is extracted from the hit structure, together with all nodes on its path which are not shared with any other path. After the hit structure has been purged, the recording of hits may continue using the space which has been released.

\end{enumerate}

\subsubsection{Noisy data}\label{ma_noisy_data_section}

Because of the way each model searches for a global optimum in the building of multiple alignments, it does not depend on the presence or absence of any particular feature or combination of features. Up to a point, plausible results may be obtained in the face of errors of omission, commission and substitution in the data. This is illustrated in the two multiple alignments in Figure \ref{noisy_data_recognition_figure} where the New pattern in row 0 of (b) is the same sentence as in (a) (`\texttt{t w o k i t t e n s p l a y}') but with the omission of the `\texttt{w}' in `\texttt{t w o}', the substitution of `\texttt{m}' for `\texttt{n}' in `\texttt{k i t t e n s}', and the addition of `\texttt{x}' within the word `\texttt{p l a y}'. Despite these errors, the best multiple alignment created by the SP model is, as shown in (b), the one that we judge intuitively to be `correct'.

\begin{figure}[!htbp]
\fontsize{06.50pt}{07.80pt}
\centering
{\bf
\begin{BVerbatim}
0                            t w o                 k i t t e n   s                   p l a y       0
                             | | |                 | | | | | |   |                   | | | |
1                            | | |          < Nr 5 k i t t e n > |                   | | | |       1
                             | | |          | |                | |                   | | | |
2                            | | |   < N Np < Nr               > s >                 | | | |       2
                             | | |   | | |                         |                 | | | |
3                   < D Dp 4 t w o > | | |                         |                 | | | |       3
                    | |            | | | |                         |                 | | | |
4              < NP < D            > < N |                         > >               | | | |       4
               | |                       |                           |               | | | |
5              | |                       |                           |        < Vr 1 p l a y >     5
               | |                       |                           |        | |            |
6              | |                       |                           | < V Vp < Vr           > >   6
               | |                       |                           | | | |                   |
7 < S Num    ; < NP                      |                           > < V |                   > > 7
       |     |                           |                                 |
8     Num PL ;                           Np                                Vp                      8

(a)

0                            t   o                 k i t t e     m s                   p l a x y       0
                             |   |                 | | | | |       |                   | | |   |
1                            |   |          < Nr 5 k i t t e n >   |                   | | |   |       1
                             |   |          | |                |   |                   | | |   |
2                            |   |   < N Np < Nr               >   s >                 | | |   |       2
                             |   |   | | |                           |                 | | |   |
3                   < D Dp 4 t w o > | | |                           |                 | | |   |       3
                    | |            | | | |                           |                 | | |   |
4              < NP < D            > < N |                           > >               | | |   |       4
               | |                       |                             |               | | |   |
5              | |                       |                             |        < Vr 1 p l a   y >     5
               | |                       |                             |        | |              |
6              | |                       |                             | < V Vp < Vr             > >   6
               | |                       |                             | | | |                     |
7 < S Num    ; < NP                      |                             > < V |                     > > 7
       |     |                           |                                   |
8     Num PL ;                           Np                                  Vp                        8

(b)
\end{BVerbatim}
}
\caption{(a) The best multiple alignment created by the SP model with a store of Old patterns like those in rows 1 to 8 (representing grammatical structures, including words) and a New pattern (representing a sentence to be parsed) shown in row 0. (b) As in (a) but with errors of omission, commission and substitution, and with same set of Old patterns as before. (a) and (b) are reproduced from Figures 1 and 2 in \protect\citet{wolff_sp_intelligent_database}, with permission.}
\label{noisy_data_recognition_figure}
\end{figure}

This kind of ability to cope gracefully with noisy data is very much in keeping with our ability to understand speech in noisy surroundings, to understand written language despite errors, and to recognise people, trees, houses, and the like, despite fog, snow, falling leaves, or other things that may obstruct our view. In a similar way, it is likely to prove useful in artificial systems for such applications as the processing of natural language and the recognition of patterns.

\subsection{Computational complexity}\label{ma_computational_complexity_section}

In considering the matching and unification of patterns, it not hard to see that, for any body of information $I$, except very small examples, there is a huge number of alternative ways in which patterns may be matched against each other, there will normally be many alternative ways in which patterns may be unified, and exhaustive search is not tractable ({\em BK}, Section 2.2.8.4).

However, with the kinds of heuristic techniques that are familiar in other AI applications---reducing the size of the search space by pruning the search tree at appropriate points, and being content with approximate solutions which are not necessarily perfect---this kind of matching becomes quite practical.\footnote{An example of how effective this rough-and-ready approach can be is the way ant colonies can find reasonably good solutions to the travelling salesman problem via the simple technique of marking their routes with pheromones and choosing routes that are most strongly marked \citep{dorigo_gambardella_1997}.} Much the same can be said about the heuristic techniques used for the building of multiple alignments (Section \ref{building_multiple_alignments_section}) and for unsupervised learning (Section \ref{ul_sp_model_section}).

For the process of building multiple alignments in the SP model, the time complexity in a serial processing environment, with conservative assumptions, has been estimated to be O$(\log_2 n \times nm)$, where $n$ is the size of the pattern from New (in bits) and $m$ is the sum of the lengths of the patterns in Old (in bits). In a parallel processing environment, the time complexity may approach O$(\log_2 n \times n)$, depending on how well the parallel processing is applied. In serial and parallel environments, the space complexity has been estimated to be O$(m)$.

Although the data sets used with the current SP model have generally been small,\footnote{Because the main focus has been on the concepts being modelled and not the speed of processing.} there is reason to be confident that the models can be scaled up to deal with large data sets because the kind of flexible matching of patterns which is at the heart of the SP model is done very fast and with huge volumes of data by all the leading internet search engines. As was suggested in Section \ref{sp_machine_section}, the relevant processes in any one of those search engines would probably provide a good basis for the creation of a high-parallel version of the SP machine.

\subsection{Calculation of probabilities associated with multiple alignments}\label{ma_probabilities_section}

As described in {\em BK} (Chapter 7), the formation of multiple alignments in the SP framework supports several kinds of probabilistic reasoning. The core idea is that any Old symbol in a multiple alignment that is {\em not} aligned with a New symbol represents an inference that may be drawn from the multiple alignment. This section outlines how probabilities for such inferences may be calculated. There is more detail in {\em BK} (Section 3.7).

\subsubsection{Absolute probabilities}\label{absolute_probabilities_section}

Any sequence of $L$ symbols, drawn from an alphabet of $|A|$ alphabetic types, represents one point in a set of $N$ points where $N$ is calculated as:

\begin{equation}
N = |A|^L.
\label{N_equation}
\end{equation}

\noindent {\em If we assume that the sequence is random or nearly so},\footnote{See {\em BK} (Section 3.7.1.1).} which means that the $N$ points are equi-probable or nearly so, the probability of any one point (which represents a sequence of length $L$) is close to:

\begin{equation}
p_{ABS} = |A|^{-L}.
\label{pABS_equation}
\end{equation}

\noindent This equation may be used to calculate the absolute probability of the code pattern that may be derived from any given multiple alignment (as described in Section \ref{ma_evaluation_section}). That number may also be regarded as the absolute probability of any inferences that may be drawn from the multiple alignment. In this calculation, $L$ is the sum of all the bits in the symbols of the code pattern and $|A|$ is 2.

As we shall see (Section \ref{generalisation_probabilities_section}), Equation \ref{pABS_equation} may, with advantage, be generalised by replacing $L$ with a value, $Lgen$, calculated in a slightly different way.

\subsubsection{Relative probabilities}\label{relative_probabilities_section}

The absolute probabilities of multiple alignments, calculated as described in the last subsection, are normally very small and not very interesting in themselves. From the standpoint of practical applications, we are normally interested in the {\em relative} values of probabilities, calculated as follows.

\begin{enumerate}

\item For the multiple alignment which has the highest $CD$ (which we shall call the {\em reference multiple alignment}), identify the {\em reference set of symbols in New}, meaning the symbols from New which are encoded by the multiple alignment.

\item Compile a {\em reference set of multiple alignments} which includes the reference multiple alignment and all other multiple alignments (if any) which encode exactly the reference set of symbols from New, neither more nor less.

\item Calculate the sum of the values for $p_{ABS}$ in the reference set of multiple alignments:

\begin{equation}
p_{A\_SUM} = \sum_{i = 1}^{i = R} p_{ABS_i}
\label{pA_SUM_equation}
\end{equation}

\noindent where $R$ is the size of the reference set of multiple alignments and $p_{ABS_i}$ is the value of $p_{ABS}$ for the $i$th multiple alignment in the reference set.

\item For each multiple alignment in the reference set, calculate its relative probability as:

\begin{equation}
p_{REL_i} = p_{ABS_i} / p_{A\_SUM}.
\label{pREL_equation}
\end{equation}

\end{enumerate}

The values of $p_{REL}$, calculated as just described, provide an effective means of comparing the multiple alignments in the reference set.

\subsubsection{A generalisation of the method for calculating absolute and relative probabilities}\label{generalisation_probabilities_section}

The value of $L$, calculated as described in Section \ref{absolute_probabilities_section}, may be regarded as the informational `cost' of encoding the New symbol or symbols that appear in the multiple alignment, excluding those New symbols that have {\em not} appeared in the multiple alignment.

This is OK but it is somewhat restrictive because it means that if we want to calculate relative probabilities for two or more multiple alignments they must all encode the same symbol or symbols from New. We cannot easily compare multiple alignments that encode different New symbols.

The generalisation proposed here is that, in the calculation of absolute probabilities, a new value, $Lgen$, would be used instead of $L$. This would be calculated as:

\begin{equation}
Lgen = L + Nnot.
\label{pABS_gen_equation}
\end{equation}

\noindent where $L$ is the total number of bits in the symbols in the code patterns (as in Section \ref{absolute_probabilities_section}) and $Nnot$ is the total number of bits in the New symbols that have {\em not} appeared in the multiple alignment.

The rationale is that, to encode {\em all} the symbols in New, we can use the code pattern to encode those New symbols that do appear in the multiple alignment and, for each of the remaining New symbols, we can simply use its code. The advantage of this scheme is that we can compare any two or more multiple alignments, regardless of the number of New symbols that appear in the multiple alignment.

\subsubsection{Relative probabilities of patterns and symbols}\label{rel_probs_patts_section}

It often happens that a given pattern from Old, or a given symbol type within patterns from Old, appears in more than one of the multiple alignments in the reference set. In cases like these, one would expect the relative probability of the pattern or symbol type to be higher than if it appeared in only one multiple alignment. To take account of this kind of situation, the SP model calculates relative probabilities for individual patterns and symbol types in the following way:

\begin{enumerate}

\item Compile a set of patterns from Old, each of which appears at least once in the reference set of multiple alignments. No single pattern from Old should appear more than once in the set.

\item For each pattern, calculate a value for its relative probability as the sum of the $p_{REL}$ values for the multiple alignments in which it appears. If a pattern appears more than once in a multiple alignment, it is only counted once for that multiple alignment.

\item Compile a set of symbol types which appear anywhere in the patterns identified in step 2.

\item For each alphabetic symbol type identified in step 3, calculate its relative probability as the sum of the relative probabilities of the patterns in which it appears. If it appears more than once in a given pattern, it is only counted once.

\end{enumerate}

The foregoing applies only to symbol types which do not appear in New. Any symbol type that appears in New necessarily has a probability of $1.0$---because it has been observed, not inferred.

\subsection{One system for both the analysis and the production of information}\label{decompression_by_compression_section}

A potentially useful feature of the SP system is that the processes which serve to analyse or parse a New pattern in terms of Old patterns, and to create an economical encoding of the New pattern, may also work in reverse, to recreate the New pattern from its encoding. This is the `output' perspective, mentioned in Section \ref{introduction_to_sp_theory_section}.

\begin{figure}[!htbp]
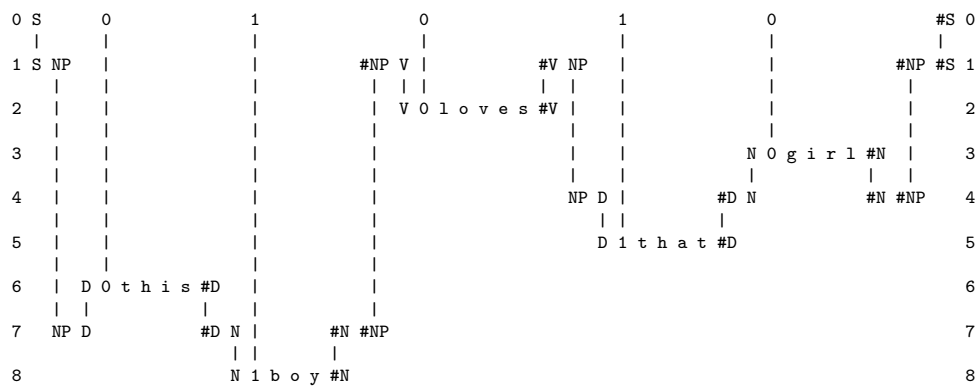

\fontsize{07.00pt}{08.40pt}
\centering
{\bf
\begin{BVerbatim}
0 S      0              1                0                   1              0                #S 0
  |      |              |                |                   |              |                |
1 S NP   |              |          #NP V |           #V NP   |              |            #NP #S 1
    |    |              |           |  | |           |  |    |              |             |
2   |    |              |           |  V 0 l o v e s #V |    |              |             |     2
    |    |              |           |                   |    |              |             |
3   |    |              |           |                   |    |            N 0 g i r l #N  |     3
    |    |              |           |                   |    |            |           |   |
4   |    |              |           |                   NP D |         #D N           #N #NP    4
    |    |              |           |                      | |         |
5   |    |              |           |                      D 1 t h a t #D                       5
    |    |              |           |
6   |  D 0 t h i s #D   |           |                                                           6
    |  |           |    |           |
7   NP D           #D N |       #N #NP                                                          7
                      | |       |
8                     N 1 b o y #N                                                              8
\end{BVerbatim}
}
\caption{The best multiple alignment found by the SP model with the New pattern `\texttt{S 0 1 0 1 0 \#S}', and the same Old patterns as were used to create the multiple alignment shown in Figure \ref{parsing_1_figure}.}
\label{parsing_2_figure}
\normalsize
\end{figure}

If the New pattern is the code sequence `\texttt{S 0 1 0 1 0 \#S}' (as described in Section \ref{multiple_alignment_section}), and if the Old patterns are the same as were used to create the multiple alignment shown in Figure \ref{parsing_1_figure}, then the best multiple alignment found by the system is the one shown in Figure \ref{parsing_2_figure}. This multiple alignment contains the same words as the original sentence (`\texttt{t h i s b o y l o v e s t h a t g i r l}'), in the same order as the original. Readers who are familiar with Prolog, will recognise that this process of recreating the original sentence from its encoding is similar in some respects to the way in which an appropriately-constructed Prolog program may be run `backwards', deriving `data' from `results'.

How is it possible to decompress the compressed code for the original sentence by using information compression? This apparent paradox---decompression by compression---may be resolved by ensuring that, when a code pattern like `\texttt{S 0 1 0 1 0 \#S}' is used to recreate the original data, each symbol is treated, at least notionally, as if contained a few more bits of information than is strictly necessary. That residual redundancy allows the system to recreate the original sentence by the same process of compression as was used to create the original parsing and encoding.\footnote{Thus `computing as compression' does not imply that all redundancy is bad and should be removed. Redundancy in information is often useful in, for example, understanding speech in noisy conditions ({\em cf.} Section \ref{ma_noisy_data_section}), or in backup copies for data.}

This process of creating a relatively large pattern from a relatively small encoding provides a model for the creation of sentences by a person or an artificial system. But instead of the New pattern being a rather dry code, like `\texttt{S 0 1 0 1 0 \#S}', it would be more plausible if it were some kind of representation of the meaning of the sentence, like that mentioned in Section \ref{ma_evaluation_section}. How a sentence may be generated from a representation of meaning is outlined in {\em BK} (Section 5.7.1).

Similar principles may apply to other kinds of `output', such as planning an outing, cooking a meal, and so on.

\section{Unsupervised learning}\label{unsupervised_learning_section}

As was mentioned in Section \ref{information_compression_section}, part of the inspiration for the SP theory has been a programme of research developing models of the unsupervised learning of language. But although the SNPR model \citep{wolff_1982} is quite successful in deriving plausible grammars from samples of English-like artificial language, it has proved to be quite unsuitable as a basis for the SP theory. In order to accommodate other aspects of intelligence, such as pattern recognition, reasoning, and problem solving, it has been necessary to develop an entirely new conceptual framework, with multiple alignment at centre stage.

So there is now the curious paradox that, while the SP theory is rooted in work on unsupervised learning, and that kind of learning has a central role in the theory, the SP model does much the same things as the earlier model, and with similar limitations (Sections \ref{unfinished_business_section} and \ref{sp_model_limitations_section}). But I believe that the new conceptual framework has many advantages, that it provides a much sounder footing for further developments, and that with some reorganisation of the learning processes in the SP computer model, its current weaknesses may be overcome (Section \ref{sp_model_limitations_section}).

\subsection{Outline of unsupervised learning in the SP model}\label{ul_sp_model_section}

The outline of the SP model in this section aims to provide sufficient detail for a good intuitive grasp of how it works. A lot more detail may be found in {\em BK} (Chapter 9).

In addition to the processes for building multiple alignments, the SP model has processes for deriving Old patterns from multiple alignments, evaluating sets of newly-created Old patterns in terms of their effectiveness for the economical encoding of the New information, and the weeding out low-scoring sets. The system does not merely record statistical information, it uses that information to learn new structures.

\subsubsection{Deriving Old patterns from multiple alignments}\label{deriving_old_patterns_section}

The process of deriving Old patterns from multiple alignments is illustrated schematically in Figure \ref{unsupervised_learning_figure}. As was mentioned in Section \ref{introduction_to_sp_theory_section}, the SP system is conceived as an abstract brain-like system that, in `input' mode, may receive `New' information via its senses and store some or all of it as `Old' information. Here, we may think of it as the brain of a baby who is listening to what people are saying. Let's imagine that he or she hears someone say `\texttt{t h a t b o y r u n s}'.\footnote{In this and other examples in this subsection, we shall assume that letters are analogues of low-level perceptual features in speech, such as formant ratios or formant transitions.} If the baby has never heard anything similar, then, if it is stored at all, that New information may be stored as a relatively straightforward copy, something like the Old pattern shown in row 1 of the multiple alignment in part (a) of the figure.

\begin{figure}[!htbp]
\fontsize{10.00pt}{12.00pt}
\centering
{\bf
\begin{BVerbatim}
0     t h a t g i r l r u n s    0
      | | | |         | | | |
1 A 1 t h a t b o y   r u n s #A 1

(a)

B 2 t h a t #B
C 3 b o y #C
C 4 g i r l #C
D 5 r u n s #D
E 6 B #B C #C D #D #E

(b)
\end{BVerbatim}
}
\caption{(a) A simple multiple alignment from which, in the SP model, Old patterns may be derived. (b) Old patterns derived from the multiple alignment shown in (a).}
\label{unsupervised_learning_figure}
\end{figure}

Now let us imagine that the information has been stored and that, at some later stage, the baby hears someone say `\texttt{t h a t g i r l r u n s}'. Then, from that New information and the previously-stored Old pattern, a multiple alignment may be created like the one shown in part (a) of Figure \ref{unsupervised_learning_figure}. And, by picking out coherent sequences that are either fully matched or not matched at all, four putative words may be extracted: `\texttt{t h a t}', `\texttt{g i r l}', `\texttt{b o y}', and `\texttt{r u n s}', as shown in the first four patterns in part (b) of the figure. In each newly-created Old pattern there are additional symbols such as `\texttt{B}', `\texttt{2}', and `\texttt{\#B}' that are added by the system, and which serve to identify the pattern, to mark its boundaries, and to mark its grammatical category or categories.

In addition to these four patterns, a fifth pattern is created, `\texttt{E 6 B \#B C \#C D \#D \#E}', as shown in the figure, that records the sequence `\texttt{t h a t~...~r u n s}', with the category `\texttt{C \#C}' in the middle representing a choice between `\texttt{b o y}' and `\texttt{g i r l}'. Part (b) in the figure is the beginnings of a grammar to describe that kind of phrase.

\subsubsection{Evaluating and selecting sets of newly-created Old patterns}\label{evaluating_and_selecting_section}

The example just described shows how Old patterns may be derived from a multiple alignment but it gives a highly misleading impression of how the SP model actually works. In practice, the program forms many multiple alignments that are much less tidy than the one shown and it creates many Old patterns that are clearly `wrong'. However, the program contains procedures for evaluating candidate sets of patterns (`grammars') and weeding out those that score badly in terms of their effectiveness for encoding the New information economically. Out of all the muddle, it can normally abstract one or two `best' grammars and these are normally ones that appear intuitively to be `correct', or nearly so. In general, the program can abstract one or more plausible grammars from a sample of English-like artificial language, including words, grammatical categories of words, and sentence structure.

In accordance with the principles of minimum length encoding \citep{solomonoff_1964}, the aim of these processes of sifting and sorting is to minimise $(G + E)$, where $G$ is the size (in bits) of the grammar that is under development and $E$ is the size (in bits) of the New patterns when they have been encoded in terms of the grammar.

For a given grammar comprising patterns $p_1 ... p_g$, the value of $G$ is calculated as:

\begin{equation}
G = \sum_{i=1}^{i=g}(\sum_{j=1}^{j=L_i}s_j)
\label{size_of_grammar_equation}
\end{equation}

\noindent where $L_i$ is the number of symbols in the $i$th pattern and $s_j$ is the encoding cost of the $j$th symbol in that pattern.

Given that each grammar is derived from a set $a_1 ... a_n$ of multiple alignments (one multiple alignment for each pattern from New), the value of $E$ for the grammar is calculated as:

\begin{equation}
E = \sum_{i=1}^{i=n}e_i
\label{size_of_encoded_data_equation}
\end{equation}

\noindent where $e_i$ is the size, in bits, of the code string derived from the $i$th multiple alignment (Section \ref{ma_evaluation_section}).

For a given set of patterns from New, a tree of alternative grammars is created with branching occurring wherever there are two or more alternative multiple alignments for a given pattern from New. The tree is grown in stages and pruned periodically to keep it within reasonable bounds. At each stage, grammars with high values for $(G + E)$ (which will be referred to as $T$) are eliminated.

\subsubsection{Plotting values for $G$, $E$ and $T$}\label{plotting_values_section}

\begin{figure}[!htbp]
\centering
\includegraphics[width=0.9\textwidth]{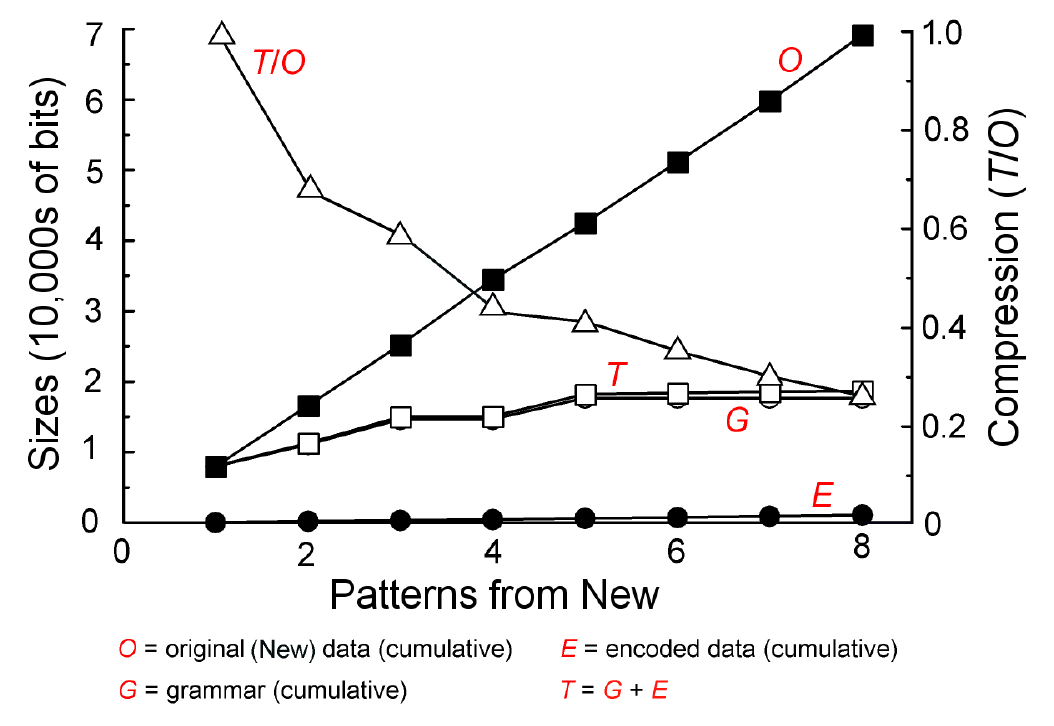}
\caption{Changing values for $G$, $E$ and $T$ and related variables as learning proceeds, as described in the text.}
\label{plotting_figure}
\end{figure}

Figure \ref{plotting_figure} shows cumulative values for $G$, $E$ and $T$ as the SP model searches for good grammars for a succession of 8 New patterns, each of which represents a sentence. Each point on each of the lower three graphs represents the relevant value (on the scale at the left) from the best grammar found after a given pattern from New has been processed. The graph labelled `$O$' shows cumulative values on the scale at the left for the succession of New patterns. The graph labelled `$T/O$' shows the amount of compression achieved (on the scale to the right).

\subsubsection{Limitations in the SP model and how they may be overcome}\label{sp_model_limitations_section}

As mentioned before (Section \ref{unfinished_business_section}), there are two main weaknesses in the processes for unsupervised learning in the SP model as it is now: the model does not learn intermediate levels in a grammar (phrases or clauses) or discontinuous dependencies of the kind described in Sections \ref{discontinuous_dependencies_section} to \ref{aux_verb_2_section}.

It appears that some reorganisation of the learning processes in the model would solve both problems. What seems to be needed is a tighter focus on the principle that, with appropriately-constructed Old patterns, multiple alignments may be created without the kind of mis-match between patterns that may be seen in Figure \ref{unsupervised_learning_figure} (a) (`\texttt{g i r l}' and `\texttt{b o y}' do not match each other), and that any such multiple alignment may be treated as if it was a simple pattern. That reform should facilitate the discovery of structures at multiple levels and the discovery of structures that are discontinuous in the sense that they can bridge intervening structures.

\subsubsection{Computational complexity}\label{learning_computational_complexity_section}

As with the building of multiple alignments (Section \ref{ma_computational_complexity_section}), the computational complexity of learning in the SP model is kept under control by pruning the search tree at appropriate points, aiming to discover grammars that are reasonably good and not necessarily perfect.

In a serial processing environment, the time complexity of learning in the SP model has been estimated to be O$(N^2)$ where $N$ is the number of patterns in New. In a parallel processing environment, the time complexity may approach O$(N)$, depending on how well the parallel processing is applied. In serial or parallel environments, the space complexity has been estimated to be O$(N)$.

\subsection{The discovery of natural structures via information compression (DONSVIC)}\label{donsvic_section}

In our dealings with the world, certain kinds of structures appear to be more prominent and useful than others: in natural languages, there are words, phrase and sentences; we understand the visual and tactile worlds to be composed of discrete `objects'; and conceptually, we recognise classes of things like `person', `house', `tree', and so on.

\begin{figure}[!htbp]
\centering
\includegraphics[width=0.5\textwidth]{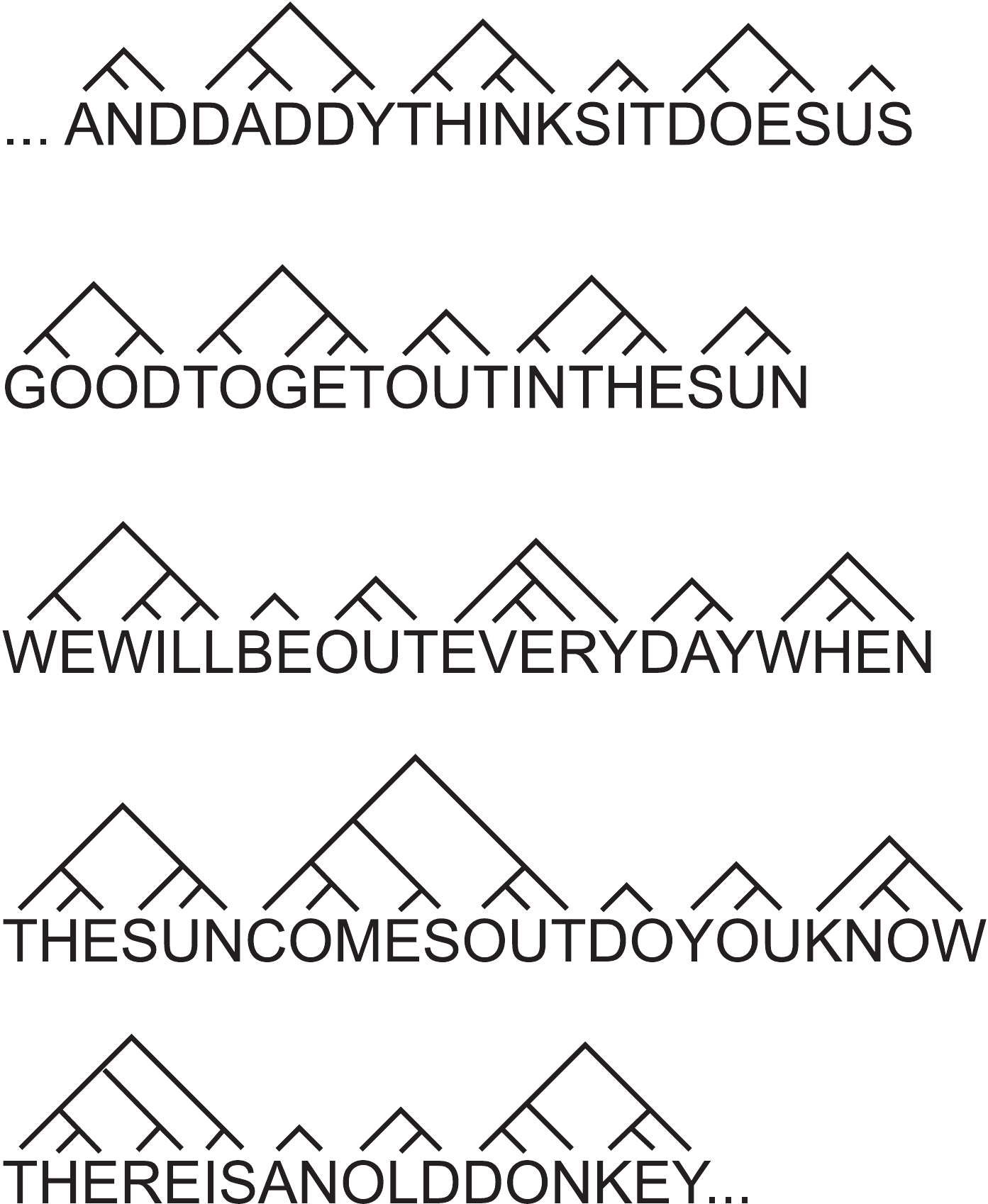}
\caption{Part of a parsing created by program MK10 \citep{wolff_1977} from a 10,000 letter sample of English (book 8A of the Ladybird Reading Series) with all spaces and punctuation removed. The program derived this parsing from the sample alone, without any prior dictionary or other knowledge of the structure of English. Reproduced from Figure 7.3 in \protect\citet{wolff_1988}, with permission.}
\label{discovery_of_words_figure}
\end{figure}

It appears that these `natural' kinds of structure are significant in our thinking because they provide a means of compressing sensory information, and that compression of information provides the key to their learning or discovery. At first sight, this looks like nonsense because popular programs for compression of information, such as those based on the LZW algorithm, or programs for JPEG compression of images, seem not to recognise anything resembling words, objects, or classes. But those programs are designed to work fast on low-powered computers. With other programs that are designed to be relatively thorough in their compression of information, natural structures can be revealed:

\begin{itemize}

\item Figure \ref{discovery_of_words_figure} shows part of a parsing of an unsegmented sample of natural language text created by the MK10 program \citep{wolff_1977} using only the information in the sample itself and without any prior dictionary or other knowledge about the structure of language. Although all spaces and punctuation had been removed from the sample, the program does reasonably well in revealing the word structure of the text. Statistical tests confirm that it performs much better than chance.

\item The same program does quite well---significantly better than chance---in revealing phrase structures in natural language texts that have been prepared, as before, without spaces or punctuation---but with each word replaced by a symbol for its grammatical category \citep{wolff_1980}. Although that replacement was done by a person trained in linguistic analysis, the discovery of phrase structure in the sample is done by the program, without assistance.

\item The SNPR program for grammar discovery \citep{wolff_1982} can, without supervision, derive a plausible grammar from an unsegmented sample of English-like artificial language, including the discovery of words, of grammatical categories of words, and the structure of sentences.

\item In a similar way, with samples of English-like artificial languages, the SP model has demonstrated an ability to learn plausible structures including words, grammatical categories of words, and the structure of sentences.

\end{itemize}

It seems likely that the principles that have been outlined in this subsection may be applied not only to the discovery of words, phrases and grammars in language-like data but also to such things as the discovery of objects in images \citep{sp_vision}, and classes of entity in all kinds of data. These principles may be characterised as {\em the discovery of natural structures via information compression}, or `DONSVIC' for short.

\subsection{Generalisation, the correction of overgeneralisations, and learning from noisy data}\label{gen_overgen_noisy_section}

\begin{figure}[!htbp]
\centering
\includegraphics[width=0.5\textwidth]{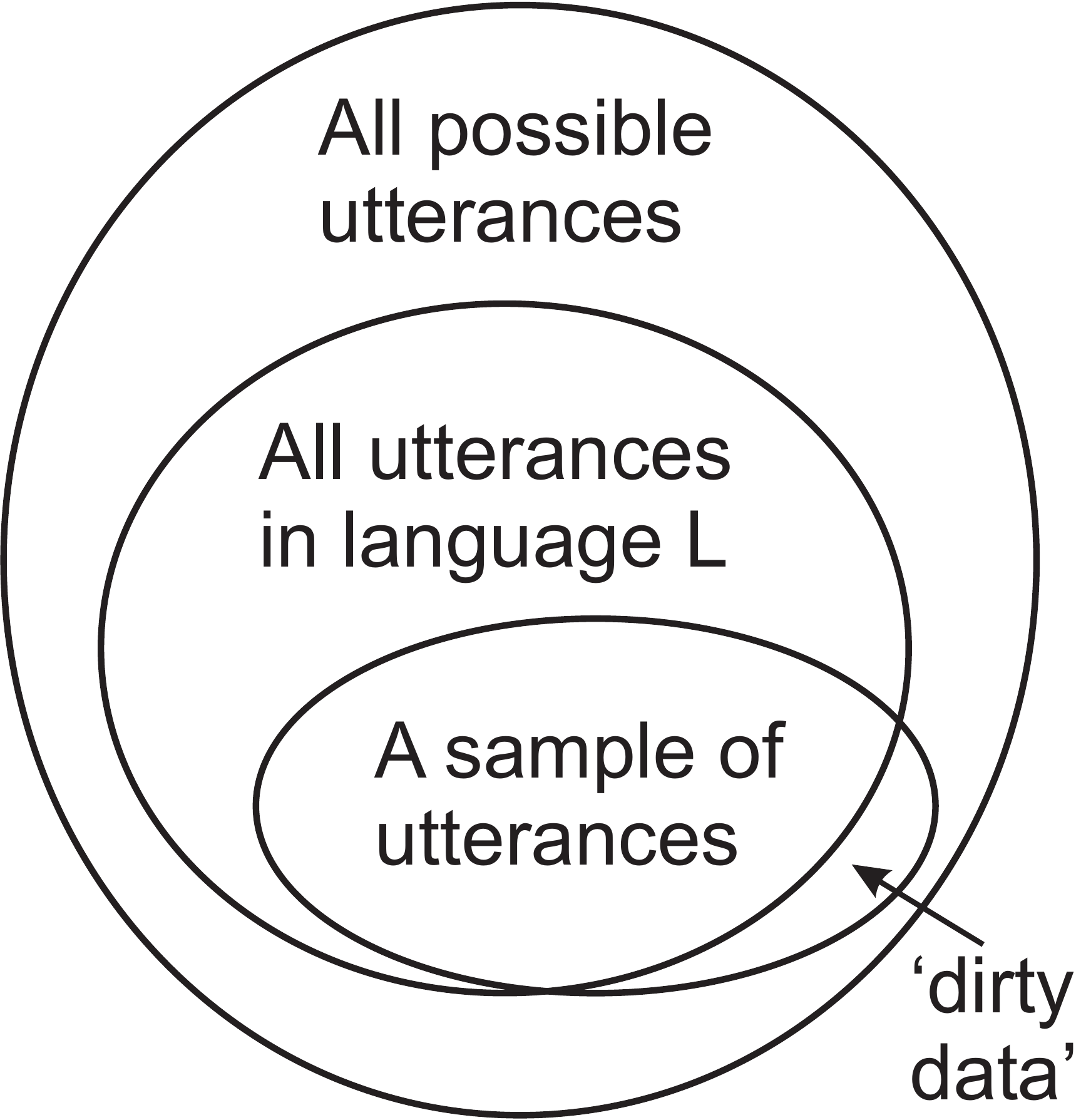}
\caption{Categories of utterances involved in the learning of a first language, $L$. In ascending order size, they are: the finite sample of utterances from which a child learns; the (infinite) set of utterances in $L$; and the (infinite) set of all possible utterances. Adapted from Figure 7.1 in \citet{wolff_1988}, with permission.}
\label{generalisation_figure}
\end{figure}

Issues that arise in the learning of a first language and, probably, in other kinds of learning, are illustrated in Figure \ref{generalisation_figure}:

\begin{itemize}

\item Given that we learn from a finite sample,\footnote{The Chomskian doctrine that children are born with a knowledge of `universal grammar' fails to account for the specifics of syntactic forms in different languages, and it depends on the still-unproven idea that there is something of substance that is shared by all the world's languages.} represented by the smallest envelope in the figure, how do we generalise from that finite sample to a knowledge of the language corresponding to the middle-sized envelope, without overgeneralising into the region between the middle envelope and the outer one?

\item How do we learn a `correct' version of our native language despite what is marked in the figure as `dirty data' (sentences that are not complete, false starts, words that are mis-pronounced, and more)?

\end{itemize}

One possible answer is that mistakes are corrected by parents, teachers, and others. But the weight of evidence is that children can learn their first language without that kind of assistance.\footnote{Relevant evident comes from cases where children learn to understand language even though they have little or no ability to speak \citep{lenneberg_1962,brown_1973}---so that there is little or nothing for anyone to correct.}

A better answer is the principle of minimum length encoding (described in its essentials in Section \ref{evaluating_and_selecting_section}):

\begin{itemize}

\item As a general rule, the greatest reductions in $(G + E)$ are achieved with grammars that represent moderate levels of generalisation, neither too little nor too much. In practice, the SNPR program, which is designed to minimise $(G + E)$, has been shown to produce plausible generalisations, without over-generalising \citep{wolff_1982}.

\item Any particular error is, by its nature, rare and so in the search for useful patterns (which, other things being equal, are the more frequently-occurring ones), it is discarded from the grammar along with other `bad' structures.\footnote{If an error is not rare it is likely to acquire the status of a dialect or idiolect variation and cease to be regarded as an error.} In the case of lossless compression, errors in any given body of data, $I$, would be retained in the encoding of $I$. But with learning, it is normally the grammar and not the encoding that is the focus of interest. In practice, the MK10 and SNPR programs have been found to be quite insensitive to errors (of omission, addition, or substitution) in their data, much as in the building of multiple alignments (Section \ref{ma_noisy_data_section}).

\end{itemize}

\subsection{One-trial learning and its implications}\label{one-trial_learning_section}

In many theories of learning,\footnote{Such as: learning in the kinds of artificial neural network that are popular in computer science; Hebb's \citeyearpar{hebb_1949} concept of learning; Pavlovian learning; and Skinnerian learning.} the process is seen as gradual: behaviour is progressively shaped by rewards or punishments or other kinds of experience.

But any theory of learning in which the process is necessarily gradual is out of step with our ordinary experience that we can and do learn things from a single experience, especially if that single experience is very significant for us ({\em BK}, Section 11.4.4.1).

In the SP theory, one-trial learning is accommodated in the way the system can store New information directly. And the gradual nature of, for example, language learning, may be explained by the complexity of the process of sifting and sorting the many alternative sets of candidate patterns to find one or more sets that are good in terms of information compression ({\em BK}, Section 11.4.4.2).

\section{Computing, mathematics, and logic}\label{computing_maths_logic_section}

Drawing mainly on {\em BK} (Chapters 4 to 11), this and the following sections describe, with a selection of examples, how the SP theory relates to several areas in artificial intelligence, mainstream computing, and human perception and cognition.

In {\em BK} (Chapter 4), I have argued that the SP system is equivalent to a universal Turing machine \citep{turing_1936}, in the sense that anything that may be computed with a Turing machine may, in principle, also be computed with an SP machine. The `in principle' qualification is necessary because the SP theory is still not fully mature and there are still some weaknesses in the SP computer models. The gist of the argument is that the operation of a Post canonical system \citep{post_1943} may be understood in terms of the SP theory and, since it is accepted that the Post canonical system is equivalent to the Turing machine (as a computational system), the Turing machine may also be understood in terms of the SP theory.

The key differences between the SP theory and earlier theories of computing are that the SP theory has a lot more to say about the nature of intelligence than earlier theories, that the theory is founded on principles of information compression via the matching and unification of patterns (`computing as compression'), and that it includes mechanisms for building multiple alignments and for heuristic search that are not present in earlier models.

\subsection{Conventional computing systems}

In conventional computing systems, compression of information may be seen in the matching of patterns with at least implicit unification of patterns that match each other---processes that appear in a variety of guises ({\em BK}, Chapter 2). And three basic techniques for the compression of information---{\em chunking-with-codes}, {\em schema-plus-correction}, and {\em run-length coding}---may be seen in various forms in the organisation of computer programs ({\em ibid}.).

\subsection{Mathematics and logic}

In a similar way, several structures and processes in mathematics and logic may be interpreted in terms of information compression via the matching and unification of patterns, and the compression techniques just mentioned ({\em BK}, Chapter 10). For example, multiplication (as repeated addition) and exponentiation (as repeated multiplication) may be seen as examples of run-length coding; a function with parameters may be seen as an example of schema-plus-correction; the chunking-with-codes technique may be seen in the organisation of number systems; and so on.

\subsection{Computing and probabilities}

As we have seen, the SP system is fundamentally probabilistic. If it is indeed Turing-equivalent, as suggested above, and if the Turing machine is regarded as a definition of `computing', then we may conclude that computing is fundamentally probabilistic. That may seem like a strange conclusion in view of the clockwork certainties that we associate with the operation of ordinary computers and the workings of mathematics and logic. There are at least three answers to that apparent contradiction:

\begin{itemize}

\item It appears that computing, mathematics and logic are more probabilistic than our ordinary experience of them might suggest. Gregory Chaitin has written: ``I have recently been able to take a further step along the path laid out by G\"{o}del and Turing. By translating a particular computer program into an algebraic equation of a type that was familiar even to the ancient Greeks, I have shown that there is randomness in the branch of pure mathematics known as number theory. My work indicates that---to borrow Einstein’s metaphor---God sometimes plays dice with whole numbers.'' \citep[p. 80]{chaitin_1988}.

\item The SP system may imitate the clockwork nature of ordinary computers by delivering probabilities of 0 and 1. This can happen with certain kinds of data, or tight constraints on the process of searching the abstract space of alternative matches, or both those things.

\item It seems likely that the all-or-nothing character of conventional computers has its origins in the low computational power of early computers. In those days, it was necessary to apply tight constraints on the process of searching for matches between patterns. Otherwise, the computational demands would have been overwhelming. Similar things may be said about the origins of mathematics and logic, which have been developed for centuries without the benefit of any computational machine, except very simple and low-powered devices. Now that it is technically feasible to apply large amounts of computational power, constraints on searching may be relaxed.

\end{itemize}

\section{Representation of knowledge}\label{representation_of_knowledge_section}

Within the multiple alignment framework (Section \ref{multiple_alignment_section}), SP patterns may serve to represent several kinds of knowledge, including grammars for natural languages,\footnote{Section \ref{nl_processing_section}; {\em BK} (Chapter 5).} ontologies,\footnote{\citet{wolff_sp_ontologies}, {\em BK} (Section 13.4.3).} class hierarchies with inheritance of attributes, including cross-classification or multiple inheritance,\footnote{{\em BK} (Section 6.4).} part-whole hierarchies and their integration with class-inclusion hierarchies,\footnote{Section \ref{part-whole_class-inclusion_section}; {\em BK} (Section 6.4).} decision networks and trees,\footnote{{\em BK} (Section 7.5).} relational tuples,\footnote{\citet{wolff_sp_intelligent_database}, {\em BK} (Section 13.4.6.1).} if-then rules,\footnote{{\em BK} {Section 7.6}.} associations of medical signs and symptoms,\footnote{\citet{wolff_medical_diagnosis}.} causal relations,\footnote{{\em BK} (Section 7.9).} and concepts in mathematics and logic such as `function', `variable', `value', `set', and `type definition'.\footnote{{\em BK} (Chapter 10).}

The use of one simple format for the representation of knowledge facilitates the seamless integration of different kinds of knowledge.

\section{Natural language processing}\label{nl_processing_section}

One of the main strengths of the SP system is in natural language processing ({\em BK}, Chapter 5):

\begin{itemize}

\item As illustrated in Figures \ref{parsing_1_figure}, \ref{noisy_data_recognition_figure} and \ref{parsing_2_figure}, grammatical rules, including words and their grammatical markers, may be represented with SP patterns.

\item Both the parsing and production of natural language may be modelled via the building of multiple alignments (Section \ref{decompression_by_compression_section}; {\em BK}, Section 5.7).

\item The system can accommodate syntactic ambiguities in language ({\em BK}, Section 5.2) and also recursive structures ({\em BK}, Section 5.3).

\item The framework provides a simple but effective means of representing discontinuous dependencies in syntax (Sections \ref{discontinuous_dependencies_section} to \ref{aux_verb_2_section}, below; {\em BK}, Sections 5.4 to 5.6).

\item The system may also model non-syntactic `semantic' structures such as class-inclusion hierarchies and part-whole hierarchies (Section \ref{part-whole_class-inclusion_section}).

\item Because there is one simple format for different kinds of knowledge, the system facilitates the seamless integration of syntax with semantics ({\em BK}, Section 5.7).

\item The system is robust in the face of errors of omission, commission or substitution in data (Sections \ref{ma_noisy_data_section} and \ref{gen_overgen_noisy_section}).

\item The importance of context in the processing of language \citep[][]{iwanska_zadrozny_1997} is accommodated in the way the system searches for a global best match for patterns: any pattern or partial pattern may be a context for any other.

\end{itemize}

\subsection{Discontinuous dependencies in syntax}\label{discontinuous_dependencies_section}

The way in which the SP system can record discontinuous dependencies in syntax may be seen in both of the two parsings in Figure \ref{noisy_data_recognition_figure}. The pattern in row 8 of each multiple alignment records the syntactic dependency between the plural noun phrase (`\texttt{t w o k i t t e n s}') which is the subject of the sentence---marked with `\texttt{Np}'---and the plural verb phrase (`\texttt{p l a y}')---marked with `\texttt{Vp}'---which belongs with it.

This kind of dependency is discontinuous because it can bridge arbitrarily large amounts of intervening structure such as, for example, `from the West' in a sentence like `Winds from the West are strong'.

This method of marking discontinuous dependencies can accommodate overlapping dependencies such as number dependencies and gender dependencies in languages like French ({\em BK}, Section 5.4). It also provides a means of encoding the interesting system of overlapping and interlocking dependencies in English auxiliary verbs, described by Noam Chomsky in {\em Syntactic Structures} \citeyearpar{chomsky_1957}.

In that book, the structure of English auxiliary verbs is part of Chomsky's evidence in support of Transformational Grammar. Despite the elegance and persuasiveness of his arguments, it turns out that the structure of English auxiliary verbs may be described with non-transformational rules in, for example, Definite Clause Grammars \citep{pereira_warren_1980}, and also in the SP system, as outlined in the subsections that follow.

\subsection{Two quasi-independent patterns of constraint in English auxiliary verbs}\label{aux_verb_1_section}

In English, the syntax for main verbs and the auxiliary verbs which may accompany them follows two quasi-independent patterns of constraint
which interact in an interesting way.

The {\em primary constraints} may be expressed with this sequence of symbols,

\begin{center}
{\bf
\begin{BVerbatim}
M H B B V,
\end{BVerbatim}
}
\end{center}

\noindent which should be interpreted in the following way:

\begin{itemize}

\item Each letter represents a category for a single word:

\begin{itemize}

\item `\texttt{M}' stands for `modal' verbs like `will', `can', `would' etc.

\item `\texttt{H}' stands for one of the various forms of the verb `to have'.

\item Each of the two instances of `\texttt{B}' stands for one of the various forms of the verb `to be'.

\item `\texttt{V}' stands for the main verb which can be any verb except a modal verb (unless the modal verb is used by itself).

\end{itemize}

\item The words occur in the order shown but any of the words may be omitted.

\item \sloppy Questions of `standard' form follow exactly the same pattern as statements except that the first verb, whatever it happens to be (`\texttt{M}', `\texttt{H}', the first `\texttt{B}', the second `\texttt{B}' or `\texttt{V}'), precedes the subject noun phrase instead of following it.

\end{itemize}

Here are two examples of the primary pattern with all of the words included:

\begin{center}
{\bf
\begin{BVerbatim}
It will have been being washed
    M    H    B     B     V

Will it have been being washed?
 M       H    B     B     V
\end{BVerbatim}
}
\end{center}

The {\em secondary constraints} are these:

\begin{itemize}

\item Apart from the modals, which always have the same form, the first verb in the sequence, whatever it happens to be (`\texttt{H}', the first `\texttt{B}', the second `\texttt{B}' or `\texttt{V}'), always has a `finite' form (the form it would take if it were used by itself with the subject).

\item If an `\texttt{M}' auxiliary verb is chosen, then whatever follows it (`\texttt{H}', first `\texttt{B}', second `\texttt{B}', or `\texttt{V}') must have an `infinitive' form (i.e., the `standard' form of the verb as it occurs in the context `to~...', but without the word `to').

\item If an `\texttt{H}' auxiliary verb is chosen, then whatever follows it (the first `\texttt{B}', the second `\texttt{B}' or `\texttt{V}') must have a past tense form such as `been', `seen', `gone', `slept', `wanted' etc. In Chomsky's {\it Syntactic Structures} \citeyearpar{chomsky_1957}, these forms were characterised as {\em en} forms and the same convention has been adopted here.

\item If the first of the two `\texttt{B}' auxiliary verbs is chosen, then whatever follows it (the second `\texttt{B}' or `\texttt{V}') must have an {\em ing} form, e.g., `singing', `eating', `having', `being' etc.

\item If the second of the two `\texttt{B}' auxiliary verbs is chosen, then whatever follows it (only the main verb is possible now) must have a past tense form (marked with {\em en}, as above).

\item The constraints apply to questions in exactly the same way as they do to statements.

\end{itemize}

Figure \ref{english_sentences_figure} shows a selection of examples with the dependencies marked.

\begin{figure}[!htbp]
\centering
{\bf
\begin{BVerbatim}
           H------en  B2---------en
          ----    --  --         --
It  will  have  been  being  washed
    ----  ----  --      ---  ----
     M----inf   B1------ing   V


          B1------ing
          --      ---
Will  he  be  talking?
----      --  ----
 M-------inf   V


              V
            ------
They  have  finished
      ----        --
       H----------en
      fin


Are  they  gone?
---        ----
B2----------en
fin         V


         B1--------ing
         --        ---
Has  he  been  working?
---        --  ----
 H---------en   V
fin
\end{BVerbatim}
}
\caption{A selection of example sentences in English with markings of dependencies between the verbs. {\em Key:} `\texttt{M}' = modal,
`\texttt{H}' = forms of the verb `have', `\texttt{B1}' = first instance of a form of the verb `be', `\texttt{B2}' = second instance of a form of the verb `be', `\texttt{V}' = main verb, `\texttt{fin}' = a finite form, `\texttt{inf}' = an infinitive form, `\texttt{en}' = a past tense form, `\texttt{ing}' = a verb ending in `ing'.}
\label{english_sentences_figure}
\end{figure}

\subsection{Multiple alignments and English auxiliary verbs}\label{aux_verb_2_section}

Without reproducing all the detail in {\em BK} (Section 5.5), we can see from Figures \ref{aux_verb_parsing_1_figure} and \ref{aux_verb_parsing_2_figure} how the primary and secondary constraints may be applied in the multiple alignment framework.

In each figure, the sentence to be analysed is shown as a New pattern in column 0. The primary constraints are applied via the matching of symbols in Old patterns in the remaining columns, with a consequent interlocking of the patterns so that they recognise sentences of the form `\texttt{M H B B V}', with options as described above.

In Figure \ref{aux_verb_parsing_1_figure},\footnote{In this figure, the sentence, `\texttt{it is wash ed}', could have been represented more elegantly as `\texttt{i t i s w a s h e d}', as in previous examples. The form shown here has been adopted because it helps to stop multiple alignments growing too large. Likewise with Figure \ref{aux_verb_parsing_2_figure}.} the secondary constraints apply as follows:

\begin{itemize}

\item The first verb, `\texttt{is}', is marked as having the finite form (with the symbol `\texttt{FIN}' in columns 5 and 7). The same word is also marked as being a form of the verb `to be' (with the symbol `\texttt{B}' in columns 4, 5 and 6). Because of its position in the parsing, we know that it is an instance of the second `\texttt{B}' in the sequence `\texttt{M H B B V}'.

\item The second verb, `\texttt{washed}', is marked as being in the {\em en} category (with the symbol `\texttt{EN}' in columns 1 and 4).

\item That a verb corresponding to the second instance of `\texttt{B}' must be followed by an {\em en} kind of verb is expressed by the pattern `\texttt{B XV EN}' in column 4.

\end{itemize}

\begin{figure}[!htbp]
\fontsize{08.00pt}{09.60pt}
\centering
{\bf
\begin{BVerbatim}
0      1      2     3    4    5     6      7      8     9     10

                                                        S
                                                        ST
                                                  NP -- NP
                                                  SNG ------- SNG
it ---------------------------------------------- it
                                                  #NP - #NP
                                           X1 --------- X1
                                           2
                                    XB1 -- XB1
                              V --- V
                         B -- B --- B
                              SNG --------------------------- SNG
                              FIN -------- FIN
                              0
is -------------------------- is
                              #V -- #V
                                    #XB1 - #XB1
                                           #X1 -------- #X1
                                           XR --------- #XR
                                           XB
                    XV - XV -------------- XV
       V ---------- V
       EN -------------- EN
       5
wash - wash
       ED --- ED
ed ---------- ed
       #ED -- #ED
       #V --------- #V
                    #S ------------------- #S --------- #S

0      1      2     3    4    5     6      7      8     9     10
\end{BVerbatim}
}
\caption{The best alignment found by the SP model with `\texttt{it is wash ed}' in New (column 0) and a user-supplied grammar in Old.}
\label{aux_verb_parsing_1_figure}
\end{figure}

In Figure \ref{aux_verb_parsing_2_figure}, the secondary constraints apply like this:

\begin{itemize}

\item The first verb `\texttt{will}' is marked as modal (with `\texttt{M}' in columns 7, 8 and 14).

\item The second verb, `\texttt{have}', is marked as having the infinitive form (with `\texttt{INF}' in columns 11 and 14) and it is also marked as a form of the verb `to have' (with `\texttt{H}' in columns 11, 12, and 15).

\item That a modal verb must be followed by a verb of infinitive form is marked with the pattern `\texttt{M INF}' in column 14.

\item The third verb, `\texttt{been}', is marked as being a form of the verb `to be' (with `\texttt{B}' in columns 2, 3 and 16). Because of its position in the parsing, we know that it is an instance of the second `\texttt{B}' in the sequence `\texttt{M H B B V}'. This verb is also marked as belonging to the {\em en} category (with `\texttt{EN}' in columns 2 and 15).

\item That an `\texttt{H}' verb must be followed by an `\texttt{EN}' verb is marked with the pattern `\texttt{H EN}' in column 15.

\item The fourth verb, `\texttt{broken}', is marked as being in the {\em en} category (with `\texttt{EN}' in columns 4 and 16).

\item That a `\texttt{B}' verb (second instance) must be followed by an `\texttt{EN}' verb is marked with the pattern `\texttt{B XV EN}' in column 16.

\end{itemize}

\begin{figure}[!htbp]
\fontsize{06.00pt}{07.20pt}
\centering
{\bf
\begin{BVerbatim}
0      1      2      3      4      5    6      7      8     9     10    11     12    13     14    15   16

                                                                  S
                                                                  Q
                                                      X1 -------- X1
                                                      0
                                               V ---- V
                                               M ---- M ----------------------------------- M
                                               0
will ----------------------------------------- will
                                               #V --- #V
                                                      #X1 ------- #X1
                                                            NP -- NP
                                                            SNG
it -------------------------------------------------------- it
                                                            #NP - #NP
                                                      XR -------- XR
                                                      XH --------------------- XH
                                                                        V ---- V
                                                                        H ---- H ---------------- H
                                                                        INF --------------- INF
have ------------------------------------------------------------------ have
                                                                        #V --- #V
                                                                               #XH
                                                      XB --------------------- XB
                                        XB ---------- XB
                     XB1 -------------- XB1
              V ---- V
              B ---- B ------------------------------------------------------------------------------- B
              EN -------------------------------------------------------------------------------- EN
be ---------- be
       EN1 -- EN1
en --- en
       #EN1 - #EN1
              #V --- #V
                     #XB1 ------------- #XB1
                                   XV - XV ---------- XV --------------------------------------------- XV
                            V ---- V
                            EN ----------------------------------------------------------------------- EN
                            1
brok ---------------------- brok
                            EN1 ---------------------------------------------------- EN1
en --------------------------------------------------------------------------------- en
                            #EN1 ----------------------------------------------------#EN1
                            #V --- #V
                                   #S - #S ---------- #S -------- #S --------- #S

0      1      2      3      4      5    6      7      8     9     10    11     12    13     14    15   16
\end{BVerbatim}
}
\caption{The best alignment found by the SP model with `\texttt{will it have be en brok en}' in New (column 0) and the same grammar in Old as was used for the example in Figure \ref{aux_verb_parsing_1_figure}.}
\label{aux_verb_parsing_2_figure}
\end{figure}

\section{Pattern recognition}\label{pattern_recognition_section}

The system also has some useful features as a framework for pattern recognition ({\em BK}, (Chapter 6):

\begin{itemize}

\item It can model pattern recognition at multiple levels of abstraction, as described in {\em BK} (Section 6.4.1), and with the integration of class-inclusion relations with part-whole hierarchies (Section \ref{part-whole_class-inclusion_section}; {\em BK}, Section 6.4.1).

\item The SP system can accommodate `family resemblance' or {\em polythetic} categories, meaning that recognition does not depend on the presence absence of any particular feature or combination of features. This is because there can be alternatives at any or all locations in a pattern, and also because of the way the system can tolerate errors in data (next point).

\item The system is robust in the face of errors of omission, commission or substitution in data (Section \ref{ma_noisy_data_section}).

\item The system facilitates the seamless integration of pattern recognition with other aspects of intelligence: reasoning, learning, problem solving, and so on.

\item A probability may be calculated for any given identification, classification, or associated inference (Section \ref{ma_probabilities_section}).

\item As in the processing of natural language (Section \ref{nl_processing_section}), the importance of context in recognition \citep[][]{oliva_torralba_2007} is accommodated in the way the system searches for a global best match for patterns. As before, any pattern or partial pattern may be a context for any other.

\end{itemize}

One area of application is medical diagnosis \citep{wolff_medical_diagnosis}, viewed as pattern recognition. There is also potential to assist in the understanding of natural vision and in the development of computer vision, as discussed in \citet{sp_vision}.

\subsection{Part-whole hierarchies, class hierarchies, and their integration}\label{part-whole_class-inclusion_section}

A strength of the multiple alignment concept is that it provides a simple but effective vehicle for the representation and processing of class-inclusion hierarchies, part-whole hierarchies, and their integration.

\begin{figure}[!htbp]
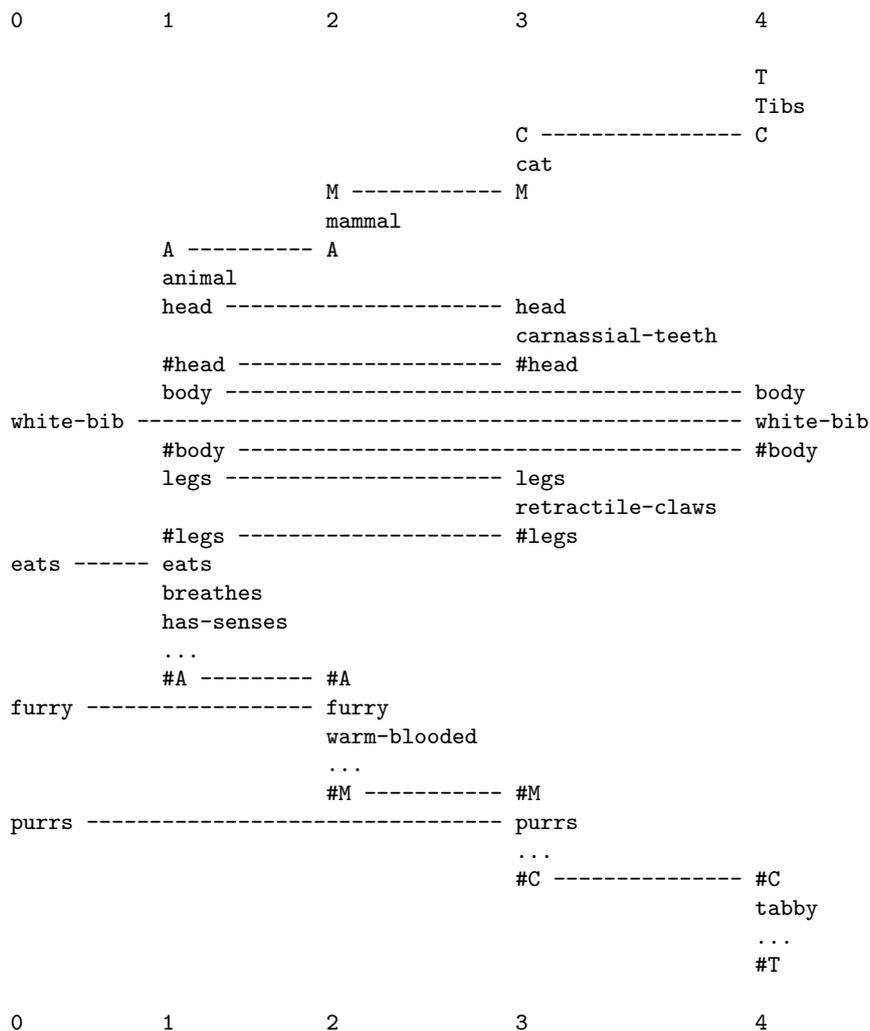

\fontsize{09.00pt}{10.80pt}
\centering
{\bf
\begin{BVerbatim}
0           1            2              3                  4

                                                           T
                                                           Tibs
                                        C ---------------- C
                                        cat
                         M ------------ M
                         mammal
            A ---------- A
            animal
            head ---------------------- head
                                        carnassial-teeth
            #head --------------------- #head
            body ----------------------------------------- body
white-bib ------------------------------------------------ white-bib
            #body ---------------------------------------- #body
            legs ---------------------- legs
                                        retractile-claws
            #legs --------------------- #legs
eats ------ eats
            breathes
            has-senses
            ...
            #A --------- #A
furry ------------------ furry
                         warm-blooded
                         ...
                         #M ----------- #M
purrs --------------------------------- purrs
                                        ...
                                        #C --------------- #C
                                                           tabby
                                                           ...
                                                           #T

0           1            2              3                  4
\end{BVerbatim}
}
\caption{The best multiple alignment found by the SP model, with the New pattern `\texttt{white-bib eats furry purrs}' and a set of Old patterns representing different categories of animal and their attributes.}
\label{class_hierarchy_figure}
\end{figure}

Figure \ref{class_hierarchy_figure} shows the best multiple alignment found by the SP model with the New pattern `\texttt{white-bib eats furry purrs}' (column 0) representing some features of an unknown creature, and with a set of Old patterns representing different classes of animal, at varying levels of abstraction. From this multiple alignment, we may conclude that the unknown entity is an animal (column 1), a mammal (column 2), a cat (column 3) and the specific individual `Tibs' (column 4).

The framework also provides for the representation of heterarchies or cross classification: a given entity, such as `Jane' (or a class of entities), may belong in two or more higher-level classes that are not themselves hierarchically related, such as `woman' and `doctor'.\footnote{Although the term `heterarchy' is not widely used, in can be useful as a means of referring to hierarchies in which, as in the example in the text, a given node may appear in two or more higher-level nodes that are not themselves hierarchically related. In the SP framework, there may be heterarchies in both class-inclusion structures and part-whole structures. But to avoid the clumsy expression `hierarchy or heterarchy', the term `hierarchy' is used in most parts of this article as a shorthand for both concepts.}

The way that class-inclusion relations and part-whole relations may be combined in one multiple alignment is illustrated in Figure \ref{class_part_plant_figure}. Here, some features of an unknown plant are expressed as a set of New patterns, shown in column 0: the plant has chlorophyll, the stem is hairy, it has yellow petals, and so on.

\begin{figure}[!htbp]
\fontsize{05.00pt}{06.00pt}
\centering
{\bf
\begin{BVerbatim}
0                 1                2                  3              4              5                  6

                  <species>
                  acris
                  <genus> ---------------------------------------------------------------------------- <genus>
                  Ranunculus ------------------------------------------------------------------------- Ranunculus
                                                                                    <family> --------- <family>
                                                                                    Ranunculaceae ---- Ranunculaceae
                                                                     <order> ------ <order>
                                                                     Ranunculales - Ranunculales
                                                      <class> ------ <class>
                                                      Angiospermae - Angiospermae
                                   <phylum> --------- <phylum>
                                   Plants ----------- Plants
                                   <feeding>
has_chlorophyll ------------------ has_chlorophyll
                                   photosynthesises
                                   <feeding>
                                   <structure> ------ <structure>
                                                      <shoot>
<stem> ---------- <stem> ---------------------------- <stem>
hairy ----------- hairy
</stem> --------- </stem> --------------------------- </stem>
                  <leaves> -------------------------- <leaves>
                  compound
                  palmately_cut
                  </leaves> ------------------------- </leaves>
                                                      <flowers> ------------------- <flowers>
                                                                                    <arrangement>
                                                                                    regular
                                                                                    all_parts_free
                                                                                    </arrangement>
                  <sepals> -------------------------------------------------------- <sepals>
                  not_reflexed
                  </sepals> ------------------------------------------------------- </sepals>
<petals> -------- <petals> -------------------------------------------------------- <petals> --------- <petals>
                                                                                    <number> --------- <number>
                                                                                                       five
                                                                                    </number> -------- </number>
                  <colour> -------------------------------------------------------- <colour>
yellow ---------- yellow
                  </colour> ------------------------------------------------------- </colour>
</petals> ------- </petals> ------------------------------------------------------- </petals> -------- </petals>
                                                                                    <hermaphrodite>
<stamens> ------------------------------------------------------------------------- <stamens>
numerous -------------------------------------------------------------------------- numerous
</stamens> ------------------------------------------------------------------------ </stamens>
                                                                                    <pistil>
                                                                                    ovary
                                                                                    style
                                                                                    stigma
                                                                                    </pistil>
                                                                                    </hermaphrodite>
                                                      </flowers> ------------------ </flowers>
                                                      </shoot>
                                                      <root>
                                                      </root>
                                   </structure> ----- </structure>
<habitat> ------- <habitat> ------ <habitat>
meadows --------- meadows
</habitat> ------ </habitat> ----- </habitat>
                  <common_name> -- <common_name>
                  Meadow
                  Buttercup
                  </common_name> - </common_name>
                                   <food_value> ----------------------------------- <food_value>
                                                                                    poisonous
                                   </food_value> ---------------------------------- </food_value>
                                   </phylum> -------- </phylum>
                                                      </class> ----- </class>
                                                                     </order> ----- </order>
                                                                                    </family> -------- </family>
                  </genus> --------------------------------------------------------------------------- </genus>
                  </species>

0                 1                2                  3              4              5                  6
\end{BVerbatim}
}
\caption{The best multiple alignment created by the SP model, with a set of New patterns (in column 0) that describe some features of an unknown plant, and a set of Old patterns, including those shown in columns 1 to 6, that describe different categories of plant, with their parts and sub-parts, and other attributes.}
\label{class_part_plant_figure}
\end{figure}

From this multiple alignment, we can see that the unknown plant is most likely to be the Meadow Buttercup, {\em Ranunculus acris}, as shown in column 1. As such, it belongs in the genus {\em Ranunculus} (column 6), the family {\em Ranunculaceae} (column 5), the order {\em Ranunculales} (column 4), the class {\em Angiospermae} (column 3), and the phylum {\em Plants} (column 2).

Each of these higher-level classifications contributes information about attributes of the plant and its division into parts and sub-parts. For example, as a member of the class {\em Angiospermae} (column 3), the plant has a shoot and roots, with the shoot divided into stem, leaves, and flowers; as a member of the family {\em Ranunculaceae} (column 5), the plant has flowers that are `regular', with all parts `free'; as a member of the phylum {\em Plants} (column 2), the buttercup has chlorophyll and creates its own food by photosynthesis; and so on.

\subsection{Inference and inheritance}\label{inheritance_of_attributes_section}

In the example just described, we can infer from the multiple alignment, very directly, that the plant which has been provisionally identified as the Meadow Buttercup performs photosynthesis (column 2), has five petals (column 6), is poisonous (column 5), and so on. As in other object-oriented systems, the first of these attributes has been `inherited' from the class `Plants', the second from the class {\em Ranunculus}, and the third from the class {\em Ranunculaceae}. These kinds of inference illustrate the close connection, often remarked, between pattern recognition and inferential reasoning \citep[see also][]{pothos_wolff_2006}.

\section{Probabilistic reasoning}\label{probabilistic_reasoning_section}

The SP system can model several kinds of reasoning including inheritance of attributes (as just described), one-step `deductive' reasoning, abductive reasoning, reasoning with probabilistic decision networks and decision trees, reasoning with `rules', nonmonotonic reasoning and reasoning with default values, reasoning in Bayesian networks (including `explaining away'), causal diagnosis, and reasoning which is not supported by evidence ({\em BK}, Chapter 7).

Since these several kinds of reasoning all flow from one computational framework (multiple alignment), they may be seen as aspects of one process, working individually or together without awkward boundaries.

Plausible lines of reasoning may be achieved, even when relevant information is incomplete.

Probabilities of inferences may be calculated, including extreme values (0 or 1) in the case of logic-like `deductions'.

A selection of examples is described in the following subsections.

\subsection{Nonmonotonic reasoning and reasoning with default values}\label{nonmonotonic_reasoning_section}

Conventional deductive reasoning is {\em monotonic} because deductions made on the strength of current knowledge cannot be invalidated by new knowledge: the conclusion that ``Socrates is mortal'', deduced from ``All humans are mortal'' and ``Socrates is human'', remains true for all time, regardless of anything we learn later. By contrast, the inference that ``Tweety can probably fly'' from the propositions that ``Most birds fly'' and ``Tweety is a bird'' is {\em nonmonotonic} because it may be changed if, for example, we learn that Tweety is a penguin.

This section presents some examples which show how the SP system can accommodate nonmonotonic reasoning.

\subsubsection{Typically, birds fly}\label{typically_birds_fly_section}

The idea that (all) birds can fly may be expressed with an SP pattern like `\texttt{Bd bird name \#name canfly warm-blooded wings feathers~...~\#Bd}'. This, of course, is an oversimplification of the real-world facts because, while it true that the majority of birds fly, we know that there are also flightless birds like ostriches, penguins and kiwis.

In order to model these facts more closely, we need to modify the pattern that describes birds to be something like this: `\texttt{Bd bird name \#name f \#f warm-blooded wings feathers~...~\#Bd}. And, to our database of Old patterns, we need to add patterns like this:

\begin{center}
{\bf
\begin{BVerbatim}
Default Bd f canfly #f #Bd #Default
P penguin Bd f cannotfly #f #Bd ... #P
O ostrich Bd f cannotfly #f #Bd ... #O.
\end{BVerbatim}
}
\end{center}

Now, the pair of symbols `\texttt{f \#f}' in `\texttt{Bd bird name \#name f \#f warm-blooded wings feathers~...~\#Bd}' functions like a `variable' that may take the value `\texttt{canfly}' if a given class of birds can fly and `\texttt{cannotfly}' when a type of bird cannot fly. The pattern `\texttt{P penguin Bd f cannotfly \#f \#Bd~...~\#P}' shows that penguins cannot fly and, likewise, the pattern `\texttt{O ostrich Bd f cannotfly \#f \#Bd~...~\#O}' shows that ostriches cannot fly. The pattern `\texttt{Default Bd f canfly \#f \#Bd \#Default}', which has a substantially higher frequency than the other two patterns, represents the default value for the variable which is `\texttt{canfly}'. Notice that all three of these patterns contains the pair of symbols `\texttt{Bd~...~\#Bd}' showing that the corresponding classes are all subclasses of birds.

\subsubsection{Tweety is a bird so, probably, Tweety can fly}\label{tweety-flies_section}

When the SP model is run with `\texttt{bird Tweety}' in New and the same patterns in Old as before, modified as just described, the three best multiple alignments found are those shown in Figures \ref{nonmon_figure_1a_figure}, \ref{nonmon_figure_1b_figure} and \ref{nonmon_figure_1c_figure}.

\begin{figure}[!htbp]
\centering
{\bf
\begin{BVerbatim}
0        1        2              3

                                 Default
                  Bd ----------- Bd
bird ------------ bird
         name --- name
Tweety - Tweety
         #name -- #name
                  f ------------ f
                                 canfly
                  #f ----------- #f
                  warm-blooded
                  wings
                  feathers
                  ...
                  #Bd ---------- #Bd
                                 #Default

0        1        2              3
\end{BVerbatim}
}
\caption{The first of the three best multiple alignments formed by the SP model with `\texttt{bird Tweety}' in New and patterns in Old as described in the text. The relative probability of this multiple alignment is 0.66.}
\label{nonmon_figure_1a_figure}
\end{figure}

\begin{figure}[!htbp]
\centering
{\bf
\begin{BVerbatim}
0        1        2              3

                                 O
                                 ostrich
                  Bd ----------- Bd
bird ------------ bird
         name --- name
Tweety - Tweety
         #name -- #name
                  f ------------ f
                                 cannotfly
                  #f ----------- #f
                  warm-blooded
                  wings
                  feathers
                  ...
                  #Bd ---------- #Bd
                                 ...
                                 #O

0        1        2              3
\end{BVerbatim}
}
\caption{The second of the three best multiple alignments formed by the SP model with `\texttt{bird Tweety}' in New and patterns in Old as described in the text. The relative probability of this multiple alignment is 0.22.}
\label{nonmon_figure_1b_figure}
\end{figure}

\begin{figure}[!htbp]
\centering
{\bf
\begin{BVerbatim}
0        1        2              3

                                 P
                                 penguin
                  Bd ----------- Bd
bird ------------ bird
         name --- name
Tweety - Tweety
         #name -- #name
                  f ------------ f
                                 cannotfly
                  #f ----------- #f
                  warm-blooded
                  wings
                  feathers
                  ...
                  #Bd ---------- #Bd
                                 ...
                                 #P

0        1        2              3
\end{BVerbatim}
}
\caption{The last of the three best multiple alignments formed by the SP model with `\texttt{bird Tweety}' in New and patterns in Old as described in the text. The relative probability of this multiple alignment is 0.12.}
\label{nonmon_figure_1c_figure}
\end{figure}

The first multiple alignment tells us that, with a relative probability of 0.66, Tweety may be the typical kind of bird that can fly. The second multiple alignment tells us that, with a relative probability of 0.22, Tweety might be an ostrich and, as such, he or she would not be able to fly. Likewise, the third multiple alignment tells us that, with a relative probability of 0.12, Tweety might be a penguin and would not be able to fly. The values for probabilities in this simple example are derived from guestimated frequencies that are, almost certainly, not ornithologically correct.

\subsubsection{Tweety is a penguin, so Tweety cannot fly}\label{tweety_is_penguin}

Figure \ref{nonmon_figure_2_figure} shows the best multiple alignment found by the SP model when it is run again, with `\texttt{penguin Tweety}' in New instead of `\texttt{bird Tweety}'. This time, there is only one multiple alignment in the reference set and its relative probability is 1.0. Correspondingly, all inferences that we can draw from this multiple alignment have a probability of 1.0. In particular, we can be confident, within the limits of the available knowledge, that Tweety cannot fly.

\begin{figure}[!htbp]
\fontsize{10.00pt}{12.00pt}
\centering
{\bf
\begin{BVerbatim}
0         1        2              3

                                  P
penguin ------------------------- penguin
                   Bd ----------- Bd
                   bird
          name --- name
Tweety -- Tweety
          #name -- #name
                   f ------------ f
                                  cannotfly
                   #f ----------- #f
                   warm-blooded
                   wings
                   feathers
                   ...
                   #Bd ---------- #Bd
                                  ...
                                  #P

0         1        2              3
\end{BVerbatim}
}
\caption{The best multiple alignment formed by the SP model with `\texttt{penguin Tweety}' in New and patterns in Old as described in the text. The relative probability of this multiple alignment is 1.0.}
\label{nonmon_figure_2_figure}
\end{figure}

In a similar way, if Tweety were an ostrich, we would be able to say with confidence (p = 1.0) that he or she would not be able to fly.

\subsection{Reasoning in Bayesian networks, including `explaining away'}\label{bayesian_networks_section}

\begin{figure}[!htbp]
\centering
\includegraphics[width=0.7\textwidth]{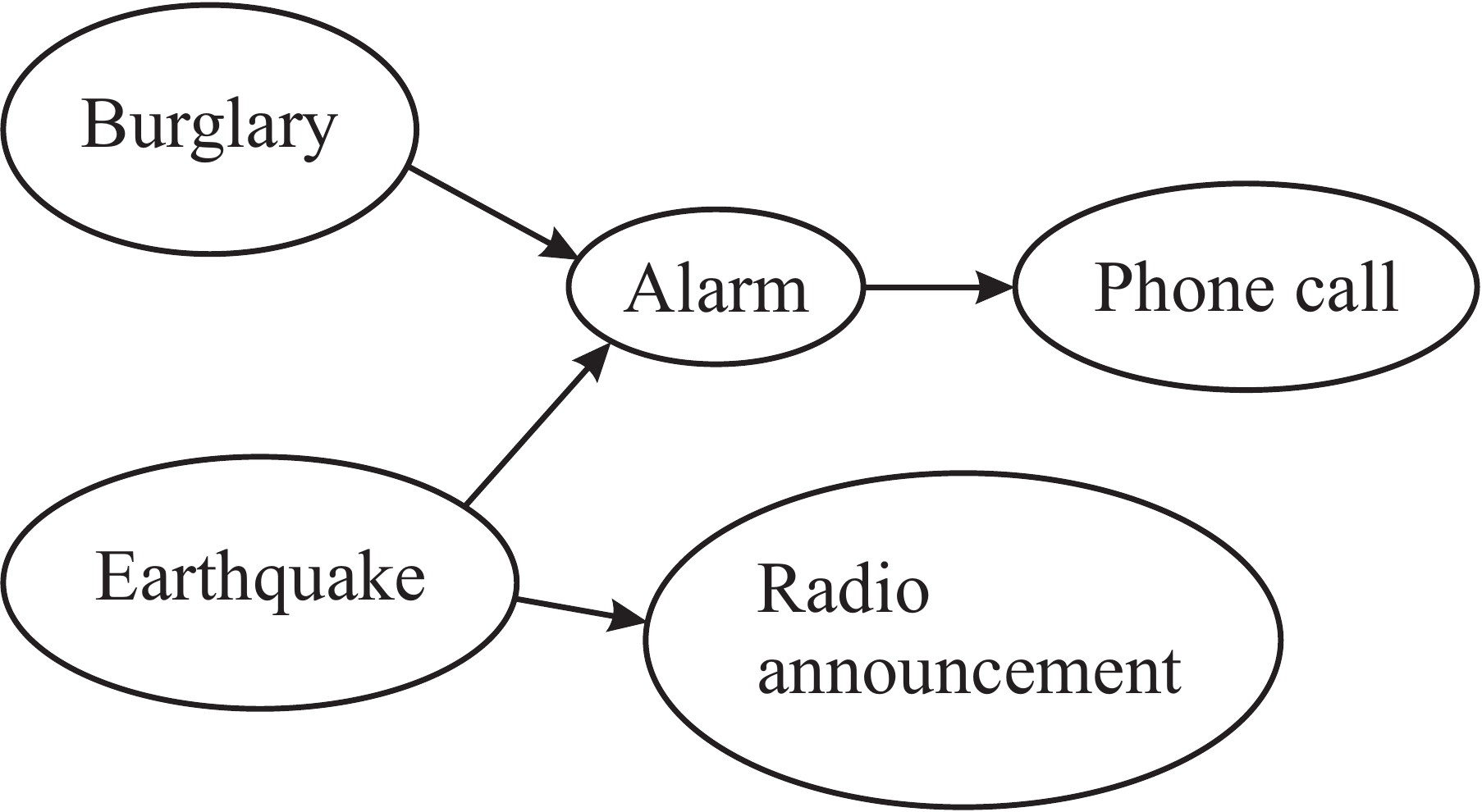}
\caption{A Bayesian network representing causal relationships discussed in the text. In this diagram, ``Phone call'' means ``a phone call about the alarm going off'' and ``Radio announcement'' means ``a radio announcement about an earthquake''.}
\label{alarm_bayesian_network_figure}
\end{figure}

A Bayesian network is a directed, acyclic graph like the one shown in Figure \ref{alarm_bayesian_network_figure}, below, where each node has zero or more `inputs' (connections with nodes that can influence the given node) and one or more `outputs' (connections to other nodes that the given node can influence).

Each node contains a set of conditional probability values, each one the probability of a given output value for a given input value or combination of input values. With this information, conditional probabilities of alternative outputs for any node may be computed for any given {\em combination} of inputs. By combining these calculations for sequences of nodes, probabilities may be propagated through the network from one or more `start' nodes to one or more `finishing' nodes.

This section describes how the SP system may perform that kind of probabilistic reasoning, and some advantages compared with Bayesian networks.

Judea Pearl \citeyearpar[p. 7]{pearl_1997} describes the phenomenon of `explaining away' like this: ``If A implies B, C implies B, and B is true, then finding that C is true makes A {\em less} credible. In other words, finding a second explanation for an item of data makes the first explanation less credible.'' (his italics). Here is an example:

\begin{quotation}

\noindent Normally an alarm sound alerts us to the possibility of a burglary. If somebody calls you at the office and tells you that your alarm went off, you will surely rush home in a hurry, even though there could be other causes for the alarm sound. If you hear a radio announcement that there was an earthquake nearby, and if the last false alarm you recall was triggered by an earthquake, then your certainty of a burglary will diminish. \citep[][pp. 8-9]{pearl_1997}.

\end{quotation}

The causal relationships in the example just described may be captured in a Bayesian network like the one shown in Figure \ref{alarm_bayesian_network_figure}.

Pearl argues that, with appropriate values for conditional probabilities, the phenomenon of ``explaining away'' can be explained in terms of this network (representing the case where there is a radio announcement of an earthquake) compared with the same network without the node for ``radio announcement'' (representing the situation where there is no radio announcement of an earthquake).

\subsubsection{Representing contingencies with patterns and frequencies}\label{contingencies_section}

To see how this phenomenon may be understood in terms of the SP theory, consider, first, the set of patterns shown in Figure \ref{alarm_patterns_figure}, which are to be stored in Old. The patterns in the figure show events which occur together in some notional sample of the `World' together with their frequencies of occurrence in the sample.

As with other knowledge-based systems, we shall assume that the `closed-world' assumption applies so that the absence of any pattern may be taken to mean that the corresponding combination of events did not occur in the period when observations were made.\footnote{Likewise, a travel booking clerk using a database of all flights between cities will assume that, if no flight is shown between, say, Edinburgh and Paris, then no such flight exists. In systems like Prolog, the closed-world assumption is the basis of `negation as failure': if a proposition cannot be proved with the clauses provided in a Prolog program then, in terms of that store of knowledge, the proposition is assumed to be false.}

\begin{figure}[!htbp]
\fontsize{10.00pt}{12.00pt}
\centering
{\bf
\begin{BVerbatim}
burglary alarm (1000)
earthquake alarm (20)
alarm phone_alarm_call (980)
earthquake radio_earthquake_announcement (40)
e1 earthquake e2 (40)
\end{BVerbatim}
}
\caption{A set of patterns to be stored in Old in an example of `explaining away'. The number in brackets after each pattern is the notional frequency of occurrence of the pattern. The symbol `\texttt{phone\_alarm\_call}' is intended to represent a phone call conveying news that the alarm sounded; `\texttt{radio\_earthquake\_announcement}' represents an announcement on the radio that there has been an earthquake. The symbols `\texttt{e1}' and `\texttt{e2}' represent other contexts for `\texttt{earthquake}' besides the contexts `\texttt{alarm}' and `\texttt{radio\_earthquake\_announcement}'.}
\label{alarm_patterns_figure}
\end{figure}

The first pattern (`\texttt{burglary alarm}') shows that there were 1000 occasions when there was a burglary and the alarm went off and the second pattern (`\texttt{earthquake alarm}') shows just 20 occasions when there was an earthquake and the alarm went off (presumably triggered by the earthquake). Thus we have assumed that, as triggers for the alarm, burglaries are much more common than earthquakes.

Since there is no pattern showing that the alarm sounded when there was a burglary and an earthquake at the same time, we may assume, via the closed-world assumption, that nothing like that happened during the sampling period.

The third pattern (`\texttt{alarm phone\_alarm\_call}') shows that, out of the 1020 cases when the alarm went off, there were 980 cases where a phone call about the alarm was made. Since there is no pattern showing phone calls about the alarm in any other context, the closed-world assumption allows us to assume that there were no false positives (eg., hoaxes)---phone calls about the alarm when no alarm had sounded.

The fourth pattern (`\texttt{earthquake radio\_earthquake\_announcement}') \newline shows that, in the sampling period, there were 40 occasions when there was an earthquake with an announcement about it on the radio. And the fifth pattern (`\texttt{e1 earthquake e2}') shows that an earthquake has occurred on 40 occasions in contexts where the alarm did not ring and there was no radio announcement.\footnote{Some of the frequencies shown in Figure \ref{alarm_patterns_figure} are intended to reflect the two probabilities suggested for this example in \citet[p. 49]{pearl_1997}: ``... the [alarm] is sensitive to earthquakes and can be accidentally (P = 0.20) triggered by one. ... if an earthquake had occurred, it surely (P = 0.40) would be on the [radio] news.''}

As before, the absence of patterns like `\texttt{earthquake alarm \newline radio\_earthquake\_announcement}' representing cases where an earthquake triggers the alarm and also leads to a radio announcement, allows us to assume via the closed-world assumption that cases of that kind have not occurred in the sampling period.

\subsubsection{Approximating the temporal order of events}\label{temporal-order_section}

In these patterns and in the multiple alignments shown below, the left-to-right order of symbols may be regarded as an approximation to the order of events in time. Thus in the first two patterns, events that can trigger an alarm precede the sounding of the alarm. Likewise, in the third pattern, `\texttt{alarm}' (meaning that the alarm has sounded) precedes `\texttt{phone\_alarm\_call}' (a phone call to say the alarm has sounded). A single dimension can only approximate the order of events in time because it cannot represent events which overlap in time or which occur simultaneously. However, this kind of approximation has little or no bearing on the points to be illustrated here.

\subsubsection{Other considerations}

Other points relating to the patterns shown in Figure \ref{alarm_patterns_figure} include:

\begin{itemize}

\item No attempt has been made to represent the idea that ``the last false alarm you recall was triggered by an earthquake'' \citep[][p. 9]{pearl_1997}. At some stage in the development of the SP system, there will be a need to take account of recency ({\em BK}, Section 13.2.6).

\item With these imaginary frequency values, it has been assumed that burglaries (with a total frequency of occurrence of 1160) are much more common than earthquakes (with a total frequency of 100). As we shall see, this difference reinforces the belief that there has been a burglary when it is known that the alarm has gone off (but without additional knowledge of an earthquake).

\item In accordance with Pearl's example (p. 49) (but contrary to the phenomenon of looting during earthquakes), it has been assumed that earthquakes and burglaries are independent. If there was some association between them, then, in accordance with the closed-world assumption, there should be a pattern in Figure \ref{alarm_patterns_figure} representing the association.

\end{itemize}

\subsubsection{Formation of alignments: the burglar alarm has sounded}\label{burglar_alarm_sounded_section}

Receiving a phone call to say that the burglar alarm at one's house has gone off may be represented by placing the symbol `\texttt{phone\_alarm\_call}' in New. Figure \ref{alarm_alignments_1_figure} shows, at the top, the best multiple alignment formed by the SP model in this case, with the patterns from Figure \ref{alarm_patterns_figure} in Old. The other two multiple alignments in the reference set are shown below the best multiple alignment, in order of CD value and relative probability. The actual values for $CD$ and relative probability are given in the caption to Figure \ref{alarm_patterns_figure}.

\begin{figure}[!htbp]
\fontsize{12.00pt}{14.40pt}
\centering
{\bf
\begin{BVerbatim}
0       phone_alarm_call 0
               |
1 alarm phone_alarm_call 1

(a)

0                phone_alarm_call 0
                        |
1          alarm phone_alarm_call 1
             |
2 burglary alarm                  2

(b)

0                  phone_alarm_call 0
                          |
1            alarm phone_alarm_call 1
               |
2 earthquake alarm                  2

(c)
\end{BVerbatim}
}
\caption{The best multiple alignment (at the top) and the other two multiple alignments in its reference set formed by the SP model with the pattern `\texttt{phone\_alarm\_call}' in New and the patterns from Figure \ref{alarm_patterns_figure} in Old. In order from the top, the values for $CD$ with relative probabilities in brackets are: 19.91 (0.656), 18.91 (0.328), and 14.52 (0.016).}
\label{alarm_alignments_1_figure}
\end{figure}

The unmatched Old symbols in these multiple alignments represent inferences made by the system. The probabilities for these inferences which are calculated by the SP model (as outlined in Section \ref{ma_probabilities_section}) are shown in Table \ref{symbol_probabilities_table}. These probabilities do not add up to 1 and we should not expect them to because any given multiple alignment may contain two or more of these symbols.

The most probable inference is the rather trivial inference that the alarm has indeed sounded. This reflects the fact that there is no pattern in Figure \ref{alarm_patterns_figure} representing false positives for telephone calls about the alarm. Apart from the inference that the alarm has sounded, the most probable inference (p = 0.328) is that there has been a burglary. However, there is a distinct possibility that there has been an earthquake---but the probability in this case (p = 0.016) is much lower than the probability of a burglary.

\begin{table}
\centering
\begin{tabular}{ll}
\em Symbol & \em Probability \\
\\
\texttt{alarm} & 1.0 \\
\texttt{burglary} & 0.328 \\
\texttt{earthquake} & 0.016 \\
\end{tabular}
\caption{The probabilities of unmatched Old symbols, calculated by the SP model for the three multiple alignments shown in Figure \ref{alarm_alignments_1_figure}.}
\label{symbol_probabilities_table}
\end{table}

These inferences and their relative probabilities seem to accord quite well with what one would naturally think following a telephone call to say that the burglar alarm at one's house has gone off (given that one was living in a part of the world where earthquakes were not vanishingly rare).

\subsection{Formation of alignments: the burglar alarm has sounded and there is a radio announcement of an earthquake}\label{radio-announcement}

In this example, the phenomenon of `explaining away' occurs when you learn not only that the burglar alarm has sounded but that there has been an announcement on the radio that there has been an earthquake. In terms of the SP model, the two events (the phone call about the alarm and the announcement about the earthquake) can be represented in New by a pattern like this:

\begin{center}
{\bf
\begin{BVerbatim}
phone_alarm_call radio_earthquake_announcement
\end{BVerbatim}
}
\end{center}

\noindent or `\texttt{radio\_earthquake\_announcement phone\_alarm\_call}'. The order of the two symbols makes no difference to the result.

\begin{figure}[!htbp]
\fontsize{09.00pt}{10.80pt}
\centering
{\bf
\begin{BVerbatim}
0                  phone_alarm_call radio_earthquake_announcement 0
                          |                       |
1            alarm phone_alarm_call               |               1
               |                                  |
2 earthquake alarm                                |               2
      |                                           |
3 earthquake                        radio_earthquake_announcement 3

(a)

0 phone_alarm_call radio_earthquake_announcement 0
                                 |
1 earthquake       radio_earthquake_announcement 1

(b)

0       phone_alarm_call radio_earthquake_announcement 0
               |
1 alarm phone_alarm_call                               1

(c)

0                phone_alarm_call radio_earthquake_announcement 0
                        |
1          alarm phone_alarm_call                               1
             |
2 burglary alarm                                                2

(d)

0                  phone_alarm_call radio_earthquake_announcement 0
                          |
1            alarm phone_alarm_call                               1
               |
2 earthquake alarm                                                2

(e)
\end{BVerbatim}
}
\caption{At the top, the best multiple alignment formed by the SP model with the pattern `\texttt{phone\_alarm\_call radio\_earthquake\_announcement}' in New and the patterns from Figure \ref{alarm_patterns_figure} in Old. Other multiple alignments formed by the SP model are shown below. From the top, the $CD$ values are: 74.64, 54.72, 19.92, 18.92, and 14.52.}
\label{alarm_alignments_2_figure}
\end{figure}

In this case, there is only one multiple alignment (shown at the top of Figure \ref{alarm_alignments_2_figure}) that can `explain' all the information in New. Since there is only this one multiple alignment in the reference set for the best multiple alignment, the associated probabilities of the inferences that can be read from the multiple alignment (`\texttt{alarm}' and `\texttt{earthquake}') are 1.0: it was an earthquake that caused the alarm to go off (and led to the phone call), and not a burglary.

These results show how `explaining away' may be explained in terms of the SP theory. The main point is that the multiple alignment or multiple alignments that provide the best `explanation' of a telephone call to say that one's burglar alarm has sounded is different from the multiple alignment or multiple alignments that best explain the same telephone call coupled with an announcement on the radio that there has been an earthquake. In the latter case, the best explanation is that the earthquake triggered the alarm. Other possible explanations have lower probabilities.

\subsubsection{Other possibilities}\label{other_possibilities_section}

As mentioned above, the closed-world assumption allows us to rule out possibilities such as:

\begin{itemize}

\item A burglary (which triggered the alarm) and, at the same time, an earthquake (which led to a radio announcement), or

\item An earthquake that triggered the alarm and led to a radio announcement and, at the same time, a burglary that did not trigger the alarm.

\item And many other unlikely possibilities of a similar kind \citep[also discussed by][Section 2.2.4]{pearl_1997}.

\end{itemize}

Nevertheless, we may consider possibilities of that kind by combining multiple alignments as described in {\em BK} (Section 7.8.7). But as a general rule, that kind of further analysis makes no difference to the original conclusion: {\em the multiple alignment which was originally judged to represent the best interpretation of the available facts has not been dislodged from this position}. This is in keeping with way we normally concentrate on the most likely explanations of events and ignore the many conceivable but unlikely alternatives.

\subsection{The SP framework and Bayesian networks}

The foregoing examples show that the SP framework is a viable alternative to Bayesian networks, at least in the kinds of situation that have been described. This subsection makes some general observations about the relative merits of the two frameworks for probabilistic reasoning where the events of interest are subject to multiple influences or chains of influence or both those things.

Without in any way diminishing Thomas Bayes' achievement, his theorem appears to have shortcomings as the basis for theorising about perception and cognition:

\begin{itemize}

\item {\em Undue complexity in the storage of statistical knowledge.} Each node in a Bayesian network contains a table of conditional probabilities for all possible combinations of inputs, and these tables can be quite large. By contrast, the SP framework only requires a single measure of frequency for each pattern. A focus on frequencies seems to yield an overall advantage in terms of simplicity compared with the representation of statistical knowledge in the form of conditional probabilities.

\item {\em Diverting attention from simpler alternatives.} By emphasising probabilities, Bayes' theorem diverts attention away from simpler and more primitive concepts of matching and unification of patterns which, by hypothesis, provide the foundation for several aspects of intelligence (Section \ref{matching_and_unification_section}).

\item {\em No place for structural learning.} Bayes' theorem assumes that the objects and categories that are to be related to each other via conditional probabilities are already `given'. It has nothing to say about how ontological knowledge may be created from raw perceptual input. By contrast, the SP framework provides for the discovery of objects and other categories via the matching and unification of patterns, in accordance with the DONSVIC principle (Section \ref{donsvic_section}).

\end{itemize}

\subsection{Causal diagnosis}\label{causal_diagnosis_section}

\begin{figure}[!htbp]
\centering
\includegraphics[width=0.7\textwidth]{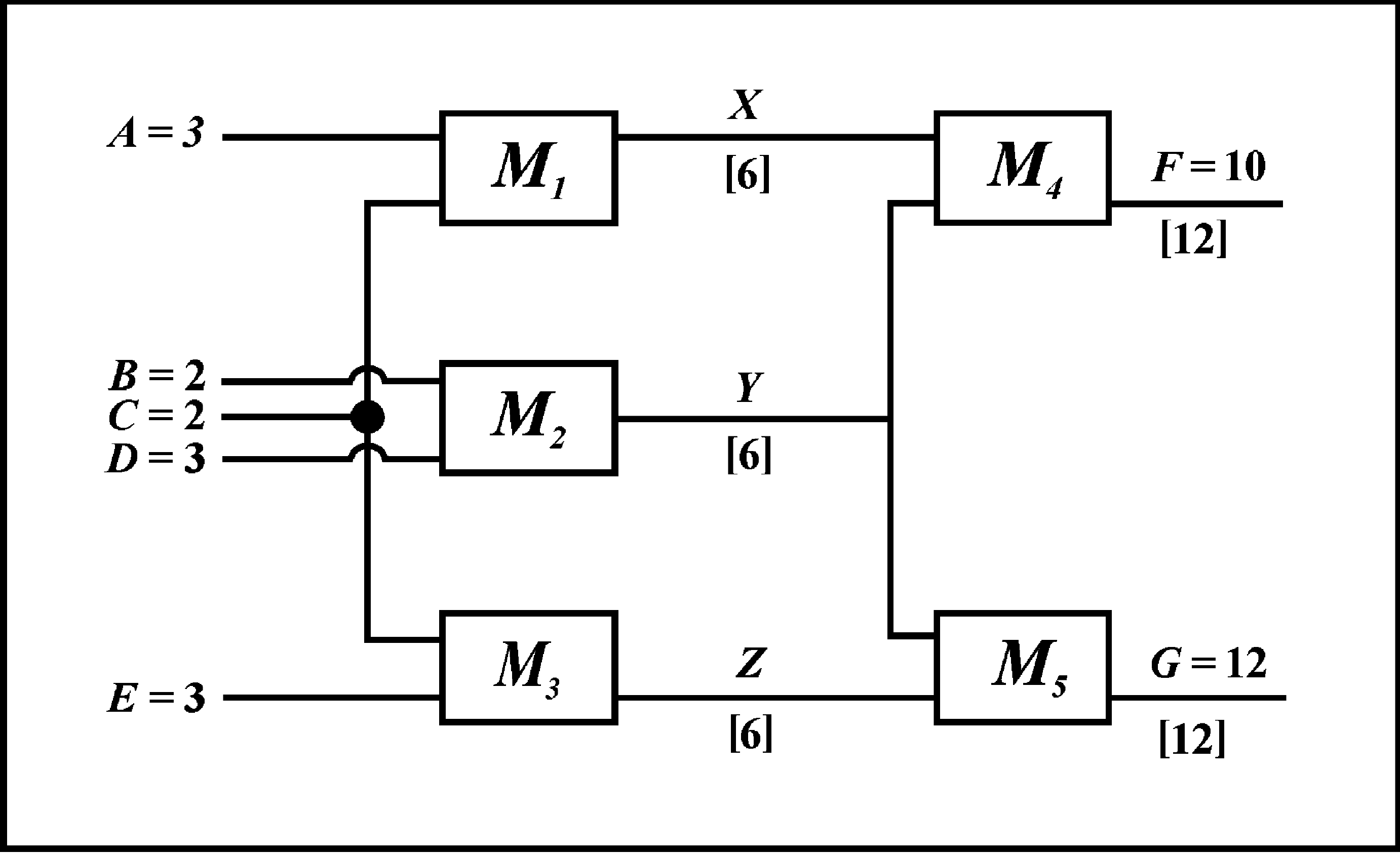}
\caption{An electronic circuit containing three multipliers, $M_1$, $M_2$, and $M_3$, and two adders, $M_4$ and $M_5$ \citep[redrawn from][p. 263]{pearl_1997}. Correct outputs are shown in square brackets.\protect\footnotemark}
\label{electronic_circuit_figure}
\end{figure}

In this section, we consider a simple example of fault diagnosis in an electronic circuit---described by \citet[pp. 263--272]{pearl_1997}. Figure \ref{electronic_circuit_figure} shows the circuit with inputs on the left, outputs on the right and, in between, three multipliers ($M_1$, $M_2$, and $M_3$) and two adders ($M_4$ and $M_5$). For the given inputs on the left, it is clear that output F is false and output G is correct.

\footnotetext{At the time of publication, permission to reproduce this diagram has been applied for twice but no reply has been received.}

\begin{figure}[!htbp]
\centering
\includegraphics[width=0.7\textwidth]{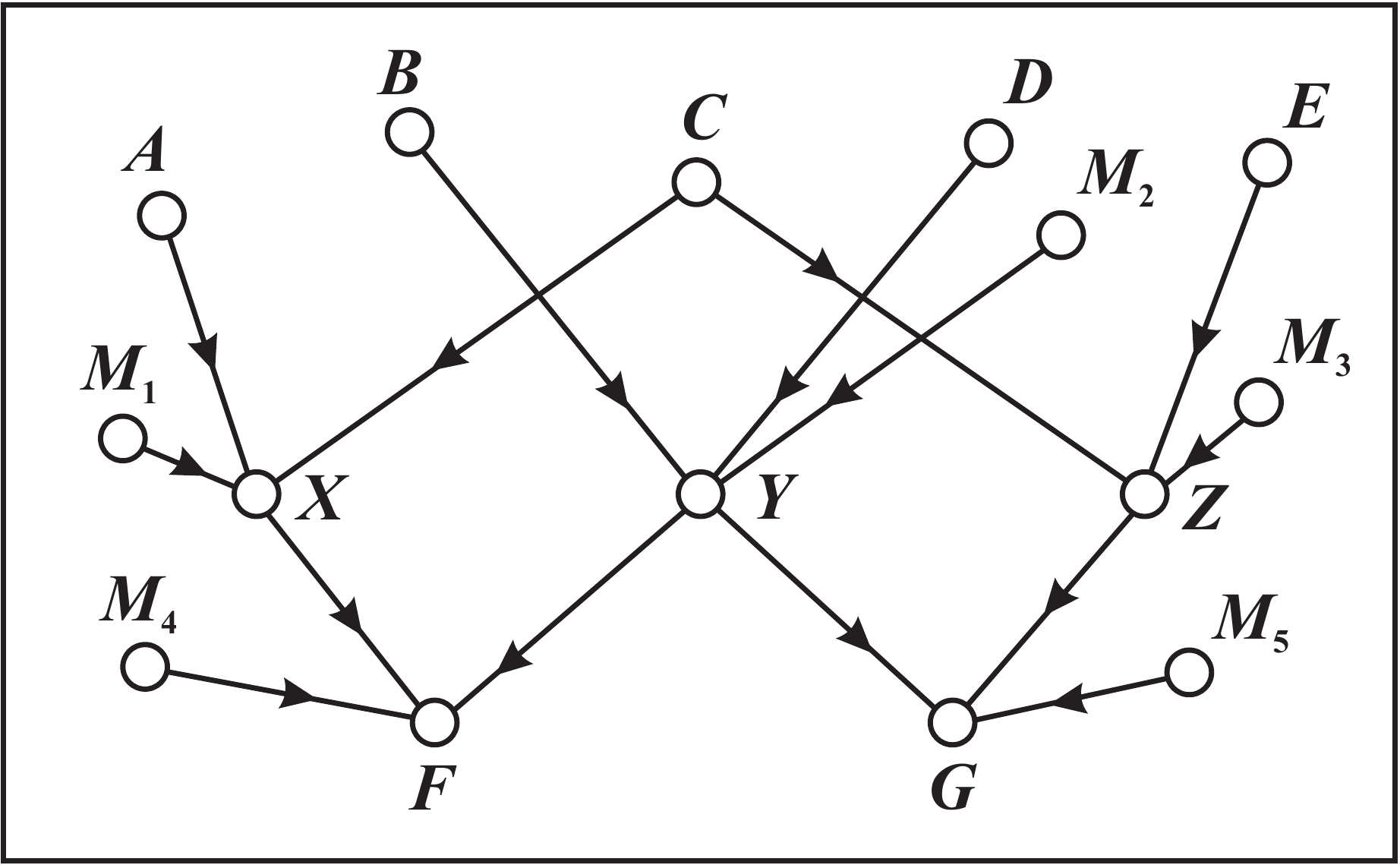}
\caption{A causal network derived from the electronic circuit in Figure \ref{electronic_circuit_figure} \citep[redrawn from][p. 264]{pearl_1997}.\protect\footnotemark}
\label{electronic_circuit_network_figure}
\end{figure}

Figure \ref{electronic_circuit_network_figure} shows a causal network derived from the electronic circuit in Figure \ref{electronic_circuit_figure} \citep[from][p. 264]{pearl_1997}. In this diagram, $X$, $Y$, $Z$, $F$ and $G$ represent the outputs of components $M_1$, $M_2$, $M_3$, $M_4$ and $M_5$, respectively. In each case, there are three causal influences on the output: the two inputs to the component and the state of the component which may be `good' or `bad'. These influences are shown in Figure \ref{electronic_circuit_network_figure} by lines with arrows connecting the source of the influence to the target node. Thus, for example, the two inputs of component $M_1$ are represented by $A$ and $C$, the good or bad state of component $M_1$ is represented by the node labelled $M_1$, and their causal influences on node $X$ are shown by the three arrows pointing at that node.

\footnotetext{At the time of publication, permission to reproduce this diagram has been applied for twice but no reply has been received.}

Given a causal analysis like this, and given appropriate information about conditional probabilities, it is possible to derive one or more alternative diagnoses of which components are good and which are bad. In Pearl's example, it is assumed that components of the same type have the same prior probability of failure and that the probability of failure of multipliers is greater than for adders. Given these assumptions and some others together with the inputs and outputs shown in Figure \ref{electronic_circuit_figure} (but not the intermediate values), the best diagnosis derived from the causal network is that the $M_1$ component is bad and the second best diagnosis is that $M_4$ is bad. Pearl indicates that some third-best interpretations may be retrievable (e.g., $M_2$ and $M_5$ are bad) ``... but in general, it is not guaranteed that interpretations beyond the second-best will be retrievable.'' (p. 272).

\subsection{An SP approach to causal diagnosis}\label{sp_causal_diagnosis}

The main elements of the SP analysis presented here are as follows:

\begin{itemize}

\item The input-output relations of any component may be represented as a set of patterns, each one with a measured or estimated frequency of occurrence.

\item With suitable extensions, these patterns may serve to transfer the output of one component to the input of another.

\item A `framework' pattern (shown at the bottom of Figure \ref{sp_causal_diagnosis_patterns_figure}) is needed to ensure that appropriate multiple alignments can be built.

\end{itemize}

Figure \ref{sp_causal_diagnosis_patterns_figure} shows a set of patterns for the circuit shown in Figure \ref{electronic_circuit_figure}. In the figure, the patterns that start with the symbol `\texttt{M1}' represent input-output relations for component $M_1$, those that start with `\texttt{M2}' represent input-output relations for the $M_2$ component and likewise for the other patterns except the last one (starting with the symbol `frame') which is the framework pattern mentioned above. For each initial symbol there is a corresponding `terminating' symbol with an initial `\texttt{\#}' character. For reasons explained shortly, there may be other symbols following the `terminating' symbol.

\begin{figure}[!htbp]
\fontsize{10.00pt}{12.00pt}
\centering
{\bf
\begin{BVerbatim}
M1 M1GOOD TM1I1 TM1I2 TM1O #M1 TM4I2 (500000)
M1 M1BAD TM1I1 TM1I2 TM1O #M1 TM4I2 (4)
M1 M1BAD TM1I1 TM1I2 FM1O #M1 FM4I2 (96)
M2 M2GOOD TM2I1 TM2I2 TM2O #M2 TM4I1 TM5I2 (500000)
M2 M2BAD TM2I1 TM2I2 TM2O #M2 TM4I1 TM5I2 (4)
M2 M2BAD TM2I1 TM2I2 FM2O #M2 FM4I1 FM5I2 (96)
M3 M3GOOD TM3I1 TM3I2 TM3O #M3 TM5I1 (500000)
M3 M3BAD TM3I1 TM3I2 TM3O #M3 TM5I1 (4)
M3 M3BAD TM3I1 TM3I2 FM3O #M3 FM5I1 (96)
M4 M4GOOD TM4I1 TM4I2 TM4O #M4 (250000)
M4 M4GOOD TM4I1 FM4I2 FM4O #M4 (250000)
M4 M4GOOD FM4I1 TM4I2 FM4O #M4 (250000)
M4 M4GOOD FM4I1 FM4I2 FM4O #M4 (250000)
M4 M4BAD TM4I1 TM4I2 FM4O #M4 (24)
M4 M4BAD TM4I1 FM4I2 FM4O #M4 (24)
M4 M4BAD FM4I1 TM4I2 FM4O #M4 (24)
M4 M4BAD FM4I1 FM4I2 FM4O #M4 (24)
M4 M4BAD TM4I1 TM4I2 TM4O #M4 (1)
M4 M4BAD TM4I1 FM4I2 TM4O #M4 (1)
M4 M4BAD FM4I1 TM4I2 TM4O #M4 (1)
M4 M4BAD FM4I1 FM4I2 TM4O #M4 (1)
M5 M5GOOD TM5I1 TM5I2 TM5O #M5 (250000)
M5 M5GOOD TM5I1 FM5I2 FM5O #M5 (250000)
M5 M5GOOD FM5I1 TM5I2 FM5O #M5 (250000)
M5 M5GOOD FM5I1 FM5I2 FM5O #M5 (250000)
M5 M5BAD TM5I1 TM5I2 FM5O #M5 (24)
M5 M5BAD TM5I1 FM5I2 FM5O #M5 (24)
M5 M5BAD FM5I1 TM5I2 FM5O #M5 (24)
M5 M5BAD FM5I1 FM5I2 FM5O #M5 (24)
M5 M5BAD TM5I1 TM5I2 TM5O #M5 (1)
M5 M5BAD TM5I1 FM5I2 TM5O #M5 (1)
M5 M5BAD FM5I1 TM5I2 TM5O #M5 (1)
M5 M5BAD FM5I1 FM5I2 TM5O #M5 (1)
frame M1 #M1 M2 #M2 M3 #M3 M4 #M4 M5 #M5 #frame (1)
\end{BVerbatim}
}
\caption{A set of SP patterns modelling I/O relations in the electronic circuit shown in Figure \ref{electronic_circuit_figure}. They were supplied as Old patterns to the SP model for the building of  the multiple alignment shown in Figure \ref{sp_causal_diagnosis_alignment_figure}. {\em Key}: `\texttt{T}' = true (information is correct); `\texttt{F}' = false (information is incorrect); `\texttt{M1}', `\texttt{M2}', `\texttt{M3}', `\texttt{M4}', `\texttt{M5}' = components of the circuit; `\texttt{GOOD}', `\texttt{BAD}' indicates whether a component is good or bad; `\texttt{I1}', `\texttt{I2}' = First and second inputs of a component; `\texttt{O}' = Output of a component.}
\label{sp_causal_diagnosis_patterns_figure}
\end{figure}

Let us now consider the first pattern in the figure (`\texttt{M1 M1GOOD TM1I1 TM1I2 TM1O \#M1 TM4I2}')
 representing I/O relations for component $M_1$ when that component is good, as indicated by the symbol `\texttt{M1GOOD}'. In this pattern, the symbols `\texttt{TM1I1}', `\texttt{TM1I2}' and `\texttt{TM1O}' represent the two inputs and the output of the component, `\texttt{\#M1}' is the terminating symbol, and `\texttt{TM4I2}' serves to transfer the output of $M_1$ to the second input of component $M_4$ as will be explained. In a symbol like `\texttt{TM1I1}', `\texttt{T}' indicates that the input is true, `\texttt{M1}' identifies the component, and `\texttt{I1}' indicates that this is the first input of the component. Other symbols may be interpreted in a similar way, following the key given in the caption of Figure \ref{sp_causal_diagnosis_patterns_figure}. In effect, this pattern says that, when the component is working correctly, true inputs yield a true output. The pattern has a relatively high frequency of occurrence (500,000) reflecting the idea that the component will normally work correctly.

The other two patterns for component $M_1$ (`\texttt{M1 M1BAD TM1I1 TM1I2 TM1O \#M1 TM4I2}' and
`\texttt{M1 M1BAD TM1I1 TM1I2 FM1O \#M1 FM4I2}') describe I/O relations when the component is bad. The first one describes the situation where true inputs to a faulty component yield a true result, a possibility noted by Pearl ({\em ibid.} p. 265). The second pattern---with a higher frequency---describes the more usual situation where true inputs to a faulty component yield a false result. Both these bad patterns have much lower frequencies than the good pattern.

The other patterns in Figure \ref{sp_causal_diagnosis_patterns_figure} may be interpreted in a similar way. Components $M_1$, $M_2$ and $M_3$ have only three patterns each because it is assumed that inputs to the circuit will always be true so it is not necessary to include patterns describing what happens when one or both of the inputs are false. By contrast, there are 4 good  patterns and 8 bad patterns for each of $M_4$ and $M_5$ because either of these components may receive faulty input.

For each of the five components, the frequencies of the bad patterns sum to 100. However, for each of the components $M_1$, $M_2$, and $M_3$, the total frequency of the good patterns is 500,000 compared with 1,000,000 for the set of good patterns associated with each of the component $M_4$ and $M_5$. These figures accord with the assumptions in Pearl's example that components of the same type have the same probability of failure and that the probability of failure of multipliers ($M_1$, $M_2$, and $M_3$) is greater than the probability of failure of adders ($M_4$ and $M_5$).

\subsection{Multiple alignments in causal diagnosis}

\begin{figure}[!htbp]
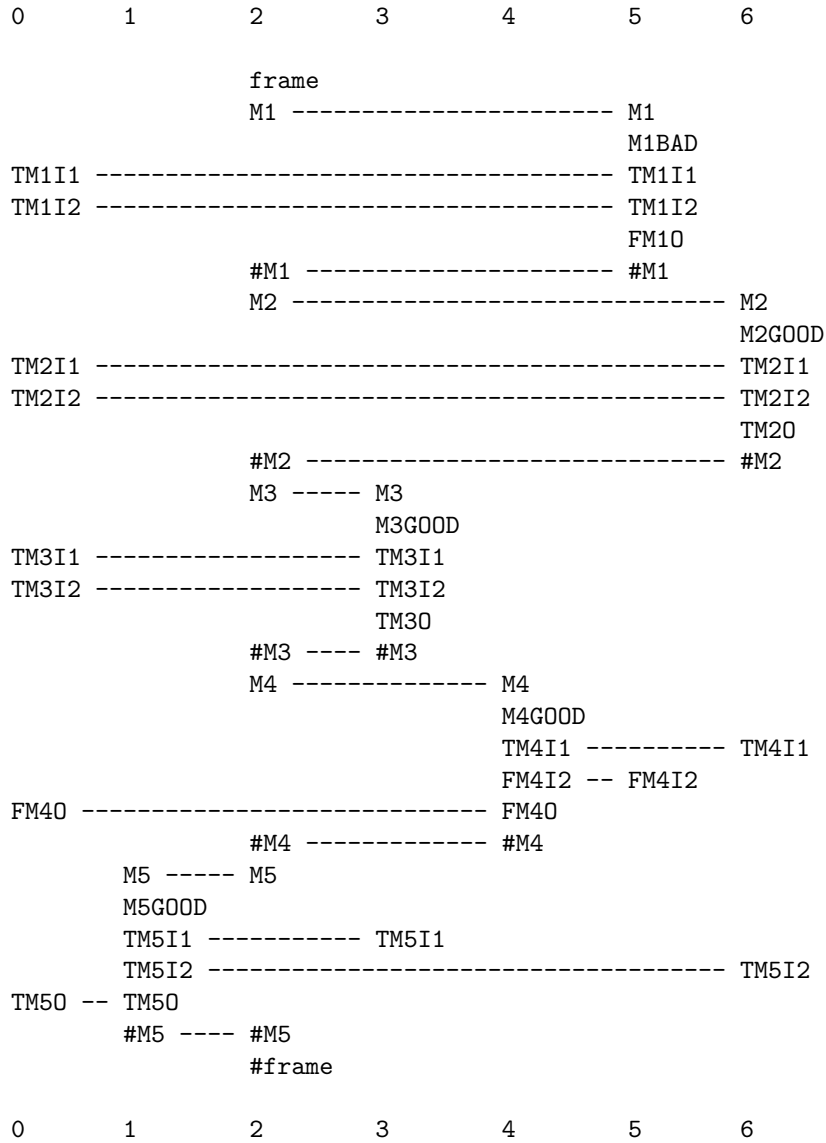

\fontsize{10.00pt}{12.00pt}
\centering
{\bf
\begin{BVerbatim}
0       1        2        3        4        5       6

                 frame
                 M1 ----------------------- M1
                                            M1BAD
TM1I1 ------------------------------------- TM1I1
TM1I2 ------------------------------------- TM1I2
                                            FM1O
                 #M1 ---------------------- #M1
                 M2 ------------------------------- M2
                                                    M2GOOD
TM2I1 --------------------------------------------- TM2I1
TM2I2 --------------------------------------------- TM2I2
                                                    TM2O
                 #M2 ------------------------------ #M2
                 M3 ----- M3
                          M3GOOD
TM3I1 ------------------- TM3I1
TM3I2 ------------------- TM3I2
                          TM3O
                 #M3 ---- #M3
                 M4 -------------- M4
                                   M4GOOD
                                   TM4I1 ---------- TM4I1
                                   FM4I2 -- FM4I2
FM4O ----------------------------- FM4O
                 #M4 ------------- #M4
        M5 ----- M5
        M5GOOD
        TM5I1 ----------- TM5I1
        TM5I2 ------------------------------------- TM5I2
TM5O -- TM5O
        #M5 ---- #M5
                 #frame

0       1        2        3        4        5       6
\end{BVerbatim}
}
\caption{The best multiple alignment found by the SP model with `\texttt{TM1I1 TM1I2 TM2I1 TM2I2 TM3I1 TM3I2 FM4O TM5O}' in New and the patterns shown in Figure \ref{sp_causal_diagnosis_patterns_figure} in Old.}
\label{sp_causal_diagnosis_alignment_figure}
\end{figure}

Given appropriate patterns, the SP model constructs multiple alignments from which diagnoses may be obtained. Figure \ref{sp_causal_diagnosis_alignment_figure} shows the best multiple alignment created by the SP model with the Old patterns shown in Figure \ref{sp_causal_diagnosis_patterns_figure} and `\texttt{TM1I1 TM1I2 TM2I1 TM2I2 TM3I1 TM3I2 FM4O TM5O}' as the New pattern. The first six symbols in this pattern express the idea that all the inputs for components $M_1$, $M_2$ and $M_3$ are true. The penultimate symbol (`FM4O') shows that the output of $M_4$ is false and the last symbol (`\texttt{TM5O}') shows that the output of $M_5$ is true---in accordance with the outputs shown in Figure \ref{electronic_circuit_figure}.

From the multiple alignment in Figure \ref{sp_causal_diagnosis_alignment_figure} it can be inferred that component $M_1$ is bad and all the other components are good. A total of seven alternative diagnoses can be derived from those multiple alignments created by the SP model that encode all the symbols in New. These diagnoses are shown in Table \ref{circuit_diagnoses_and_probabilities}, each with its relative probability.

\begin{table}
\centering
\begin{tabular}{ll}
\em Bad Component(s) & \em Relative Probability \\
\\
\texttt{M1} & 0.6664 \\
\texttt{M4} & 0.3332 \\
\texttt{M1}, \texttt{M3} & 0.00013 \\
\texttt{M1}, \texttt{M2} & 0.00013 \\
\texttt{M1}, \texttt{M4} & 6.664e-5 \\
\texttt{M3}, \texttt{M4} & 6.664e-5 \\
\texttt{M1}, \texttt{M2}, \texttt{M3} & 2.666e-8 \\
\end{tabular}
\caption{Seven alternative diagnoses of faults in the circuit shown in Figure \ref{electronic_circuit_figure}, derived from multiple alignments created by the SP model with `\texttt{TM1I1 TM1I2 TM2I1 TM2I2 TM3I1 TM3I2 FM4O TM5O}' in New and the patterns from Figure \ref{sp_causal_diagnosis_patterns_figure} in Old. The relative probability of each diagnosis is shown in the second column.}
\label{circuit_diagnoses_and_probabilities}
\end{table}

It is interesting to see that the best diagnosis derived by the SP model ($M_1$ is bad) and the second best diagnosis ($M_4$ is bad) are in accordance with first two diagnoses obtained by Pearl's method. The remaining five diagnoses derived by the SP model are different from the one obtained by Pearl's method ($M_2$ and $M_5$ are bad) but this is not altogether surprising because detailed frequencies or probabilities are different from Pearl's example and there are differences in assumptions that have been made.

\section{Information storage and retrieval}\label{info_storage_retrieval_section}

The SP theory provides a versatile model for database systems, with the ability to accommodate object-oriented structures, as well as relational `tuples', and network and tree models of data \citep{wolff_sp_intelligent_database}. It lends itself most directly to information retrieval in the manner of query-by-example but it appears to have potential to support the use of natural language or query languages such as SQL.

Unlike some ordinary database systems:

\begin{itemize}

\item The storage and retrieval of information is integrated with other aspects of intelligence such as pattern recognition, reasoning, planning, problem solving, and learning---as outlined elsewhere in this article.

\item The SP system provides a simple but effective means of combining class hierarchies with part-whole hierarchies, with inheritance of attributes (Section \ref{part-whole_class-inclusion_section}).

\item It provides for cross-classification with multiple inheritance.

\item There is flexibility and versatility in the representation of knowledge arising from the fact that the system does not distinguish `parts' and `attributes' \citep[][Section 4.2.1]{wolff_sp_intelligent_database}.

\item Likewise, the absence of a distinction between `class' and `object' facilitates the representation of knowledge and eliminates the need for a `metaclass' \citep[][Section 4.2.2]{wolff_sp_intelligent_database}.

\item SP patterns provide a simpler and more direct means of representing entity-relationship models than do relational tuples \citep[][Section 4.2.3]{wolff_sp_intelligent_database}.

\end{itemize}

\section{Planning and problem solving}\label{planning_problem_solving_section}

The SP framework provides a means of planning a route between two places, and, with the translation of geometric patterns into textual form, it can solve the kind of geometric analogy problem that may be seen in some puzzle books and IQ tests ({\em BK}, Chapter 8).

\begin{figure}[!htbp]
\centering
\includegraphics[width=0.7\textwidth]{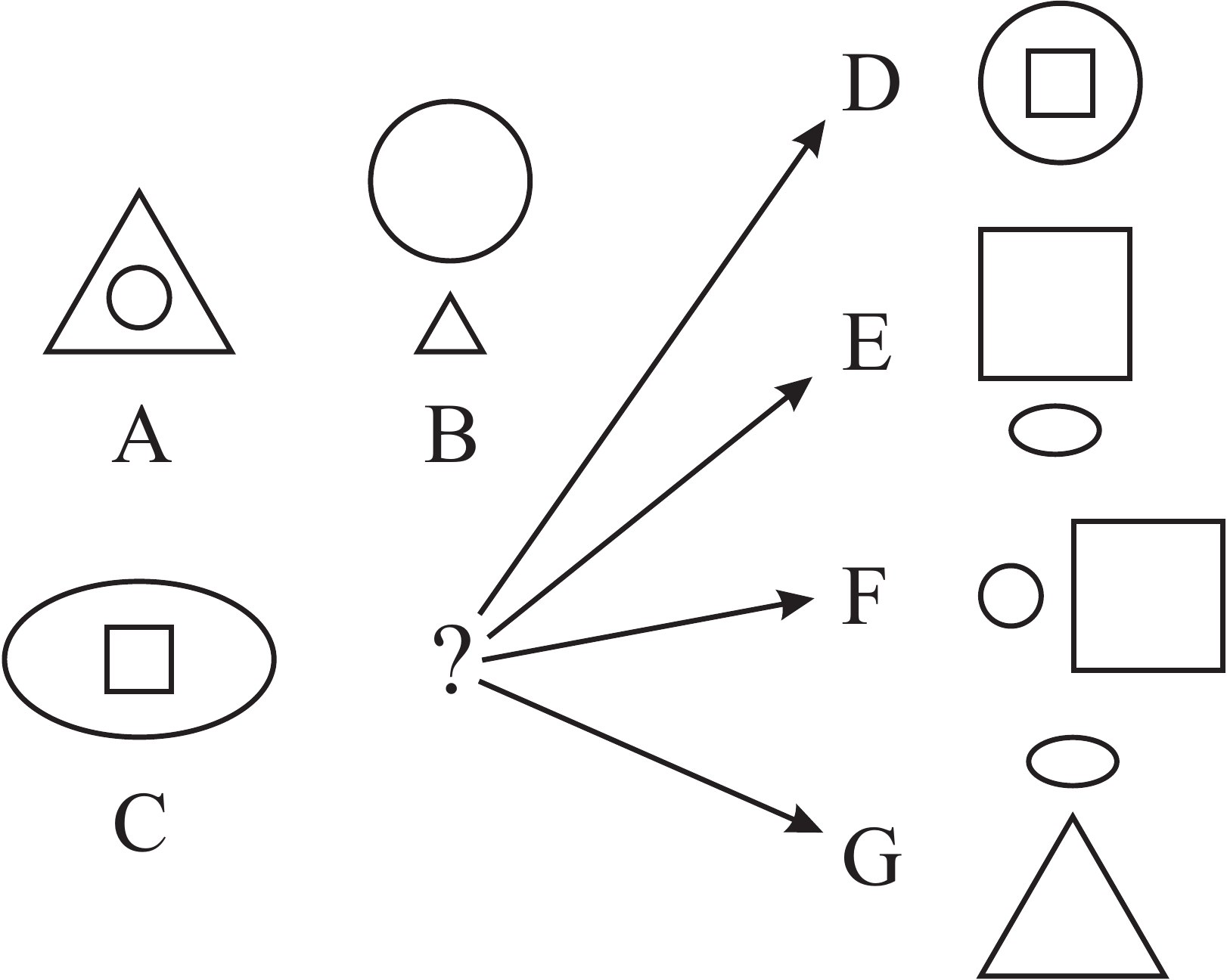}
\caption{A geometric analogy problem.}
\label{geometric_analogy_figure}
\end{figure}

Figure \ref{geometric_analogy_figure} shows an example of the latter kind of problem. The task is to complete the relationship ``A is to B as C is to ?'' using one of the figures `D', `E', `F' or `G' in the position marked with `?'. For this example, the `correct' answer is clearly `E'. Quote marks have been used for the word `correct' because in many problems of this type, there may be two or even more alternative answers for which cases can be made and there is a corresponding uncertainty about which answer is the right one.

Computer-based methods for solving this kind of problem have existed for some time (e.g., Evans' \citeyearpar{evans_1968} well-known heuristic algorithm). In more recent work \citep{belloti_gammerman_1996, gammerman_1991}, minimum length encoding principles have been applied to good effect. This kind of problem may also be understood in terms of the SP concepts.

As in most previous work, the proposed solution assumes that some mechanism is available which can translate the geometric forms in each problem into patterns of text symbols like other patterns in this article. For example, item `A' in Figure \ref{geometric_analogy_figure} may be described as `\texttt{small circle inside large triangle}'.

How this kind of translation may be done is not part of the present proposals (one such translation mechanism is described in \citet{evans_1968}). As noted elsewhere \citep{gammerman_1991}, successful solutions for this kind of problem require consistency in the way the translation is done. For this example, it would be unhelpful if item `A' in Figure \ref{geometric_analogy_figure} were described as `\texttt{large triangle outside small circle}' while item `C' were described as `\texttt{small square inside large ellipse}'. For any one puzzle, the description needs to stick to one or other of `X outside Y' or `Y inside X'---and likewise for `above/below' and `left-of/right-of'.

Given that the diagrammatic form of the problem has been translated into patterns as just described, this kind of problem can be cast as a problem of partial matching, well within the scope of the SP model. To do this, symbolic representations of item A and item B in Figure \ref{geometric_analogy_figure} are treated as a single pattern, thus:

\begin{center}
{\bf
\begin{BVerbatim}
small circle inside large triangle ;
     large circle above small triangle
\end{BVerbatim}
}
\end{center}

\noindent and this pattern is placed in New. Four other patterns are constructed by pairing a symbolic representation of item C (on the left) with symbolic representations of each of D, E, F and G (on the right), thus:

\begin{center}
{\bf
\begin{BVerbatim}
C1 small square inside large ellipse ;
     D small square inside large circle #C1
C2 small square inside large ellipse ;
     E large square above small ellipse #C2
C3 small square inside large ellipse ;
     F small circle left-of large square #C3
C4 small square inside large ellipse ;
     G small ellipse above large rectangle #C4.
\end{BVerbatim}
}
\end{center}

\noindent These four patterns are placed in Old, each with an arbitrary frequency value of 1.

Figure \ref{geometric_analogy_alignment_figure} shows the best multiple alignment found by the SP model with New and Old as just described. The multiple alignment is a partial match between the New pattern (in column 0) and the second of the four patterns in Old (in column 1). This corresponds with the `correct' result (item E) as noted above.

\begin{figure}[!htbp]
\fontsize{12.00pt}{14.40pt}
\centering
{\bf
\begin{BVerbatim}
0          1

           C2
small ---- small
circle     square
inside --- inside
large ---- large
triangle   ellipse
; -------- ;
           E
large ---- large
circle     square
above ---- above
small ---- small
triangle   ellipse
           #C2

0          1
\end{BVerbatim}
}
\caption{The best multiple alignment found by the SP model for the patterns in New and Old as described in the text.}
\label{geometric_analogy_alignment_figure}
\end{figure}

\section{Compression of information}\label{compression_potential_section}

Since information compression is central to the workings of the SP system, it is natural to consider whether the system might provide useful insights in that area. In that connection, the most promising aspects of the SP system appear to be:

\begin{itemize}

\item The discovery of recurrent patterns in data via the building of multiple alignments, with heuristic search to sift out the patterns that are most useful in terms of compression.

\item The potential of the system to detect and encode discontinuous dependencies in data. It appears that there is potential here to extract kinds of redundancy in information that are not accessible via standard methods for the compression of information.

\end{itemize}

In terms of the trade-off that exists between computational resources that are required and the level of compression that can be achieved, it is intended that the system will operate towards the `up market' end of the spectrum---by contrast with LZW algorithms and the like, which have been designed to be `quick-and-dirty', sacrificing performance for speed on low-powered computers.

\section{Perception, cognition and neuroscience}\label{perception_cognition_neuroscience_section}

Since much of the inspiration for the SP theory has come from evidence, mentioned in Section \ref{information_compression_section}, that, to a large extent, the workings of brains and nervous systems may be understood in terms of information compression, the theory is about perception and cognition as well as artificial intelligence and mainstream computing.

That said, the main elements of the theory---the multiple alignment concept in particular---are theoretical constructs derived from what appears to be necessary to model, in an economical way, such things as pattern recognition, reasoning, and so on. In {\em BK} (Chapter 12), there is some discussion of how the SP concepts relate to a selection of issues in human perception and cognition. A particular interest at the time of writing (after that chapter was written) is the way that the SP theory may provide an alternative to quantum probability as an explanation of phenomena such as the `conjunction fallacy' \cite[see, for example,][]{pothos_busemeyer_2013}.

\begin{figure}[!htbp]
\centering
\includegraphics[width=0.6\textwidth]{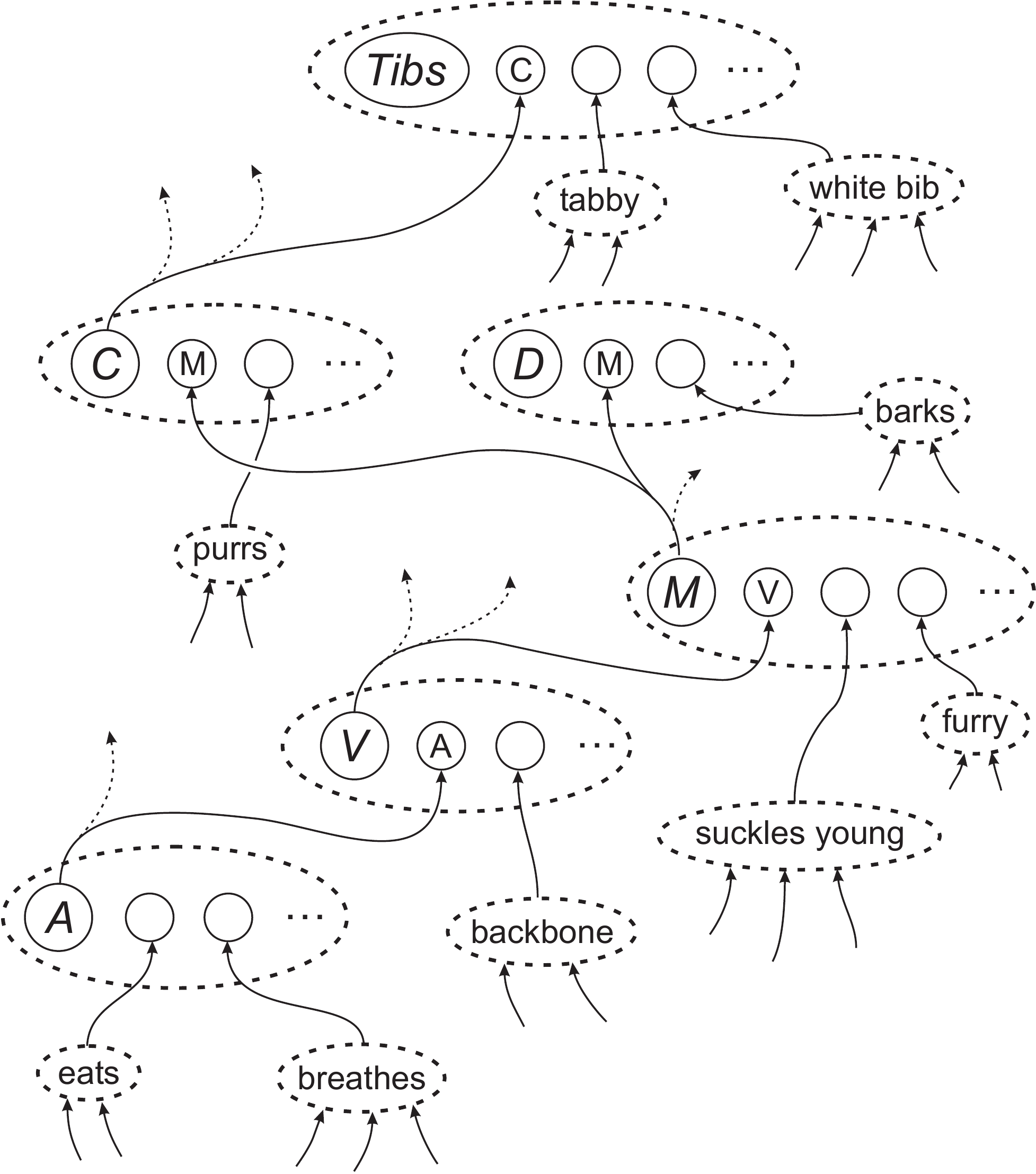}
\caption{Schematic representation of hypothesised neural analogues of SP patterns and their inter-connections. {\em Key}: `C' = cat, `D' = dog, `M' = mammal, `V' = vertebrate, `A' = animal, `...' = further structure that would be shown in a more comprehensive example. Pattern assemblies are surrounded by broken lines and each neuron is represented by an unbroken circle or ellipse. Lines with arrows show connections between pattern assemblies and the flow of sensory signals. Connections between neurons within each pattern assembly are not marked.}
\label{neural_analogue_figure}
\end{figure}

In {\em BK} (Chapter 11), I have described in outline, and tentatively, how such things as SP patterns and multiple alignments may be realised with neurons and connections between them. The cortex of the brains of mammals---which is, topologically, a two-dimensional sheet---may be, in some respects, like a sheet of paper on which {\em pattern assemblies} may be written. These are neural analogues of SP patterns, shown schematically in Figure \ref{neural_analogue_figure}. Unlike information written on a sheet of paper, there are neural connections between patterns---as shown in the figure---and communications amongst them.

These proposals, which are adapted with modifications from Hebb's \citeyearpar{hebb_1949} concept of a `cell assembly', are very different from how artificial `neural networks' are generally conceived in computer science.\footnote{See, for example, ``Artificial neural network'', {\em Wikipedia}, \href{http://en.wikipedia.org/wiki/Artificial\_neural\_network}{en.wikipedia.org/wiki/Artificial\_neural\_network}, retrieved 2013-05-10.} As noted in Section \ref{one-trial_learning_section}, learning in the SP system is very different from learning in that kind of network---or Hebbian learning.

\section{Conclusion}

The SP theory aims to simplify and integrate concepts across artificial intelligence, mainstream computing and human perception and cognition, with information compression as a unifying theme. The matching and unification of patterns and the concept of multiple alignment are central ideas.

In accordance with Occam's Razor, the SP system combines conceptual simplicity with descriptive and explanatory power. A relatively simple mechanism provides an interpretation for a range of concepts and phenomena in several areas including conepts of `computing', aspects of mathematics and logic, representation of knowledge, natural language processing, pattern recognition, several kinds of probabilistic reasoning, information storage and retrieval, planning and problem solving, information compression, neuroscience, and human perception and cognition.

As suggested in Section \ref{sp_machine_section}, an aid to further research would be the creation of a high-parallel, open-source version of the SP machine, that may be accessed via the web.

% \bibliography{latex_references}

\end{document}